\documentclass[twoside]{article}

\usepackage[accepted]{aistats2024}
\usepackage[round]{natbib}
\usepackage{float}

\usepackage{amssymb}
\usepackage{bm}
\usepackage{svg}

\usepackage{booktabs, multirow}
\usepackage{soul}
\usepackage{subfiles} %
\usepackage{subcaption} 
\usepackage{graphicx}
\usepackage{ifthen}
\usepackage{makecell}
\usepackage{adjustbox}
\usepackage{tabularx}

\usepackage{authblk}

\usepackage{wrapfig}

\usepackage[toc,page]{appendix}
\usepackage{tocloft}
\usepackage{titletoc}

\usepackage{hyperref}
\hypersetup{
    colorlinks=true,
    linkcolor=blue,
    filecolor=magenta,      
    urlcolor=cyan
}

\newcommand{\name}{Mixture-of-Linear-Experts}
\newcommand{\nameshort}{MoLE}
\newcommand{\papertitle}{Mixture-of-Linear-Experts for Long-term Time Series Forecasting}

\begin{document}

% If your paper is accepted and the title of your paper is very long,
% the style will print as headings an error message. Use the following
% command to supply a shorter title of your paper so that it can be
% used as headings.
%
%\runningtitle{I use this title instead because the last one was very long}

% If your paper is accepted and the number of authors is large, the
% style will print as headings an error message. Use the following
% command to supply a shorter version of the authors names so that
% they can be used as headings (for example, use only the surnames)
%
%\runningauthor{Surname 1, Surname 2, Surname 3, ...., Surname n}

\twocolumn[

\aistatstitle{\papertitle}

\aistatsauthor{ Ronghao Ni\textsuperscript{1} \And Zinan Lin\textsuperscript{2} \And Shuaiqi Wang\textsuperscript{1} \And Giulia Fanti\textsuperscript{1} }
\vspace{10pt}
\aistatsaddress{ $^1$Carnegie Mellon University \And  $^2$Microsoft Research \AND \{ronghaon, shuaiqiw, gfanti\}@andrew.cmu.edu \And zinanlin@microsoft.com}]

\fancyhead[CO]{\small\bfseries Ronghao Ni, Zinan Lin, Shuaiqi Wang, Giulia Fanti}

\begin{abstract}
    Long-term time series forecasting (LTSF) aims to predict future values of a time series given the past values. 
    The current state-of-the-art (SOTA) on this problem is attained in some cases by \emph{linear-centric} models, which primarily feature a linear mapping layer. 
    However, due to their inherent simplicity, they are not able to adapt their prediction rules to periodic changes in time series patterns. 
    To address this challenge, we propose a Mixture-of-Experts-style augmentation for linear-centric models and propose \name{} (\nameshort{}).
    Instead of training a single model, \nameshort{} trains multiple linear-centric models (i.e., experts) and a router model that weighs and mixes their outputs.
    While the entire framework is trained end-to-end, each expert learns to specialize in a specific temporal pattern, and the router model learns to compose the experts adaptively.
    Experiments show that \nameshort{} reduces forecasting error of  linear-centric models, including DLinear, RLinear, and RMLP, in over {78}\% of the datasets and settings we evaluated. By using \nameshort{} existing linear-centric models can achieve SOTA LTSF results in 68\% of the experiments that PatchTST reports and we compare to, whereas existing single-head linear-centric models  achieve SOTA results in only {25}\% of cases.
    % Additionally, \nameshort{} models achieve SOTA in all settings for the newly released Weather2K datasets.
\end{abstract}

\section{INTRODUCTION}

Long-term time series forecasting (LTSF) is an important problem in the machine learning community, given its application in areas like weather modeling \citep{zhu2023weather2k, wu_interpretable_2023}, traffic flow prediction \citep{pdformer}, and financial forecasting \citep{ariyo2014stock}. 

Various classical methods \citep{box1976time,fildes1991forecasting,han2019review} and deep-learning methods \citep{bai2018empirical, borovykh2017conditional,lai2018modeling, chang2018memory, fan2019multi} have been used for this task, including transformers \citep{zhou2021informer,wu2021autoformer,zhang2022crossformer,nie2022time}. 
However, recent studies show that in some settings,
\emph{linear-centric models} surpass prior baselines---including transformers---sometimes by a significant margin \citep{zeng2023transformers,li2023revisiting}. 
Linear-centric models feature a single linear layer, possibly combined with non-linear pre- and postprocessing steps. 
Examples include DLinear, NLinear \citep{zeng2023transformers}, RLinear, and RMLP \citep{li2023revisiting}. 

% However, real-world time series are usually \emph{non-stationary}.
However, real-world time series often exhibit \emph{seasonal variations} and \emph{non-stationarity}. 
For example, traffic patterns change on different days of the week. Due to the inherent simplicity of linear-centric models, it is difficult for them to capture these patterns.

In this paper, we propose \name{} (\nameshort{}) to address the above limitation. %
We propose to train multiple linear-centric models (i.e., experts) to
collaboratively predict the time series. 
A router model, 
which accepts a timestamp embedding of the input sequence as input, learns to weigh these experts adaptively. This layer is supposed to learn the periodicity of the time series and adjust the weights for each expert accordingly, ensuring that different experts specialize in different periods of the time series. 
Note that \nameshort{} can be applied to any linear-centric model as-is.  

More broadly, Mixture-of-Experts (MoE) has a long history \citep{jordan1994hierarchical,6215056} and was recently revived due to its successful application in SOTA language \citep{artetxe2021efficient,yi2023edgemoe,shen2023flan} and image models \citep{riquelme2021scaling,dryden2022spatial}. Its main motivation in these domains has been \emph{efficiency}: it increases model capacity without much inference overhead by training multiple experts simultaneously and only activating \emph{a subset of} experts for a given input (a.k.a. Sparse Mixture-of-Experts \citep{shazeer2017outrageously}). In contrast, we show that MoE over linear-centric models gives significant gains in forecasting \emph{fidelity}; \nameshort{} keeps the simplicity of linear-centric models while using  MoE to model diverse temporal patterns. 

Our primary contributions include:
\begin{itemize}
\item We propose \nameshort{}, a mixture-of-experts approach that can augment any linear-centric LTSF model to capture non-stationary, seasonal temporal patterns. %
\item We demonstrate through comprehensive empirical evaluations that \nameshort{} improves upon existing linear-centric methods, including those that are currently SOTA for the LTSF problem. 
{Concretely, we observe when \nameshort{} was applied to DLinear, it enhanced its performance in {32}/44 experimental settings ({73}\% of experiments). Similarly, with the RLinear and RMLP models, integrating \nameshort{} resulted in improvements in {38}/44 scenarios ({86}\%) and {33}/44 scenarios ({75}\%) respectively.}
We find that among the 28 datasets and settings that PatchTST reports and we compare to, 
\nameshort{} allows linear-centric models to achieve SOTA results in 19 cases (68\%), whereas without \nameshort{} (single-head), they are SOTA in {7} settings ({25}\%). 
% For the Weather2K datasets, \nameshort{}-linear-centric models achieve SOTA in all settings.
\item We conduct careful ablation studies to explain the reasons for the success of \nameshort{}; these show that the observed performance boost is due to the %
specialized time-aware experts
rather than just an expanded model size. Moreover, the gains of \nameshort{} are more pronounced when input sequence lengths are short compared to the prediction sequence length.
\end{itemize}

In all, \nameshort{} offers an efficient and simple enhancement for linear-centric LTSF models, which can be used out-of-the-box for improved forecasting. 

\section{RELATED WORK}
We highlight three categories of LSTF models: those based on transformers, linear-centric models, and models based on other architectures. 

\paragraph{Transformer-based Models} Transformers have shown great potential and excellent performance in long-term time series forecasting \citep{zhou2021informer, wu2021autoformer, zhang2022crossformer, wen2022transformers}. The first well-known transformer for LTSF, Informer \citep{zhou2021informer}, addresses issues like quadratic time complexity with ProbSparse self-attention and a generative style decoder. Following Informer, models such as Autoformer \citep{wu2021autoformer}, Pyraformer \citep{liu2021pyraformer}, Preformer \citep{du2023preformer}, and FEDFormer \citep{zhou2022fedformer} were introduced. Autoformer \citep{wu2021autoformer} uses decomposition and auto-correlation for performance, Pyraformer \citep{liu2021pyraformer} focuses on multiresolution attention for signal processing efficiency, Preformer \citep{du2023preformer} introduces segment-wise correlation for efficient attention calculation, and FEDFormer \citep{zhou2022fedformer} combines frequency analysis with Transformers for enhanced time series representation. The state-of-the-art model, PatchTST \citep{nie2022time}, shifts focus to the importance of patches, enhancing the model's capability to capture local and global dependencies in data. We compare against PatchTST \citep{nie2022time} as a SOTA transformer-based baseline.

\paragraph{Linear-centric Models}

Despite the success of transformer-based models in LTSF, \cite{zeng2023transformers} raised doubts about their efficacy. They proposed three linear or linear-centric models: vanilla linear, DLinear, and NLinear, which outperform existing transformer-based models by a large margin. Among the three models, DLinear is most commonly compared to in subsequent work \citep{nie2022time, li2023revisiting}. We include DLinear as a baseline model that can be integrated with MoLE.

Following the work of \cite{zeng2023transformers}, several other linear-centric models have been proposed \citep{li2023revisiting, xu2023fits}. We compare against  RLinear and RMLP \citep{li2023revisiting} as the only linear-centric models that outperform PatchTST \citep{nie2022time} on some datasets and settings.

\paragraph{Other Architectures}

Other common LTSF models are based on multi-layer perceptrons (MLP) \citep{
wang2023st, zhou2022treedrnet, zhang2022less, shao2022spatial, chen2023tsmixer, das2023long} and convolutional neural networks (CNN) \citep{
wang2022micn, wang2023tlnets, gong2023patchmixer}. Since these models are not directly evaluated in papers focusing on linear-centric and transformer-based models, we have not included them in our evaluation. The main point of this work is not to compare linear-centric models to transformers, but to propose a technique that improves linear centric-models. 

\cite{gruver2023large} recently introduced the LLM\textsc{Time} model, which employs large language models as a zero-shot predictor for time series. We were unable to compare with this model due to the prohibitive costs of API calls to proprietary LLMs. Additionally, their quantitative results were presented as bar charts without exact numbers, preventing a direct comparison of reported outcomes.

\section{PRELIMINARIES}

\subsection{Problem Definition}

\paragraph{Time Series Forecasting} Given an input sequence $\mathbf{X} \in \mathbb{R}^{c\times s}$, where $c$ represents the number of channels (features) and $s$ denotes the number of historical (input) timestamps, our objective is to find a mapping $f(\mathbf{X})=\mathbf{Y}$ such that $\mathbf{Y}\in \mathbb{R}^{c\times p}$ and $p$ represents the number of future (output) timestamps to predict. Additionally, we are provided with the dates and times corresponding to each timestamp of $\mathbf{X}$ and $\mathbf{Y}$, represented as $\bm{x}_{mark} \in \mathbb{R}^s$ and $\bm{y}_{mark} \in \mathbb{R}^p$ respectively. We assume timestamps are regularly-spaced, meaning for all $i, j \in [0, s+p)$, the relationship $\bm{xy}_{mark}^{(i+1)} - \bm{xy}_{mark}^{(i)} = \bm{xy}_{mark}^{(j+1)} - \bm{xy}_{mark}^{(j)}$ holds, where $\bm{xy}_{mark}$ represents the stacking of $\bm{x}_{mark}$ and $\bm{y}_{mark}$. 
Our goal is to determine the mapping $f$ that minimizes a specific loss function $loss(\mathbf{Y}, \mathbf{\hat{Y}})$, where $\mathbf{\hat{Y}} \in \mathbb{R}^{c\times p}$ denotes the ground truth.

\paragraph{Long-term Time Series Forecasting} Long-term Time Series Forecasting (LTSF) involves predicting far into the future. Although there is no standard criterion to differentiate between long- and short-term forecasting, LTSF experiments often set the minimum prediction length (i.e., $p$) to 96.

\subsection{Linear-Centric LTSF Models}

\begin{figure}[t!]
\vspace{.2in}
\centerline{\includegraphics[width=0.5\textwidth]{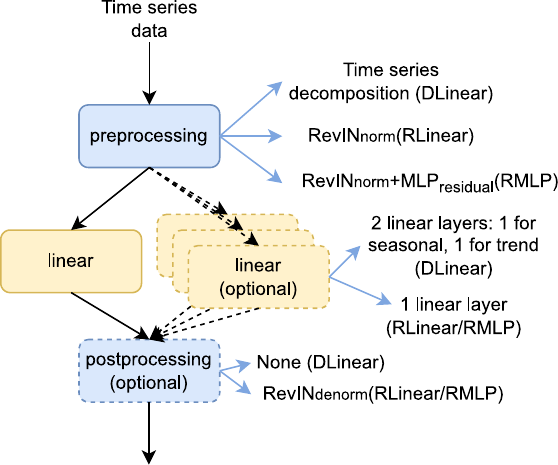}}
\vspace{.3in}
\caption{A common structure of linear-centric LTSF models. 
}
\label{fig:common_linear}
\end{figure}

\cite{zeng2023transformers} introduced a set of linear-centric models, LTSF Linears, which are recognized as the first linear-centric models to challenge the promising performance of transformers in time series forecasting. Among these, DLinear performs the best and is now commonly used as a baseline. Building on this work, \cite{li2023revisiting} looked further into the state-of-the-art transformer for LTSF, PatchTST \citep{nie2022time}, and concluded that the linear mapping, reversible normalization (RevIN) \citep{kim2021reversible}, and channel independence (CI) are important mechanisms in the model. By applying only these components to simple linear models, RLinear and RMLP outperformed PatchTST across many datasets, and are currently SOTA.
We next explain these architectures, which are illustrated in Figure \ref{fig:common_linear}.

\paragraph{DLinear} DLinear \citep{zeng2023transformers} fuses the decomposition layer from Autoformer \citep{wu2021autoformer} with a simple linear layer as follows: A moving average kernel initially decomposes the input data into two parts, termed trend and seasonal (the remainder after applying the moving average). Each of these segments then goes through a 1-layer linear layer. The outputs are summed up to produce the final predictions. Precisely, this can be written as $\mathbf{Y} = \mathbf{W}_{trend}\mathbf{X}_{trend} +  \mathbf{W}_{seasonal}\mathbf{X}_{seasonal}$, where $\mathbf{X}_{trend} \in \mathbb{R}^{c\times s}$ and $\mathbf{X}_{seasonal} \in \mathbb{R}^{c\times s}$ denote the decomposed inputs. $\mathbf{W}_{trend} \in \mathbb{R}^{s\times p}$ and $\mathbf{W}_{seasonal} \in \mathbb{R}^{s\times p}$ represent two linear layers along the temporal axis, and $\mathbf{Y} \in \mathbb{R}^{c\times p}$ is the prediction.

\paragraph{RLinear} RLinear \citep{li2023revisiting}  combines RevIN  \citep{kim2021reversible} with a single-layer linear layer. RevIN is a simple yet effective normalization technique that consists of a normalization and denormalization process, combined with a learnable affine transformation. This method is designed to adapt to distribution shifts for more accurate time-series forecasting. Notably, unlike DLinear, the RevIN used in RLinear has trainable parameters, which guide the affine transformation of the input. After the linear layer, the output is then transformed back using the same set of parameters and the statistics calculated from the input data. This process can be described as $\mathbf{Y} = \operatorname{RevIN_{denorm}}(\mathbf{W} \operatorname{RevIN_{norm}}(\mathbf{X}))$ where $\operatorname{RevIN_{denorm}} = \operatorname{RevIN_{norm}}^{-1}$, and $\mathbf{W} \in \mathbb{R}^{s \times p}$ is the linear layer.

\paragraph{RMLP} RMLP \citep{li2023revisiting} incorporates an additional 2-layer MLP to the base RLinear model. The 2-layer MLP operates within a residual block before the original linear layer. The updated model can be expressed as $\mathbf{Y} = \operatorname{RevIN_{denorm}}(\mathbf{W} (\mathbf{X}_{norm} + \operatorname{MLP}(\mathbf{X}_{norm})))$ where $\mathbf{X}_{norm} = \operatorname{RevIN_{norm}}(\mathbf{X})$ and $\operatorname{MLP}$ is the newly added 2-layer MLP. This model was introduced to achieve better performance on some larger-scale datasets where RLinear does not perform adequately.

\section{OUR METHOD: \name{} (\nameshort{})}
\label{sec:method}

We next introduce our proposed method: \name{} (\nameshort{}). 
We use the shorthand \nameshort{}-X (e.g., \nameshort{}-DLinear) to refer to a linear-centric model X  that has been augmented with \nameshort{}.

\subsection{Model Architecture}

\begin{figure}[t]
\vspace{.2in}
\centerline{\includegraphics[width=0.5\textwidth]{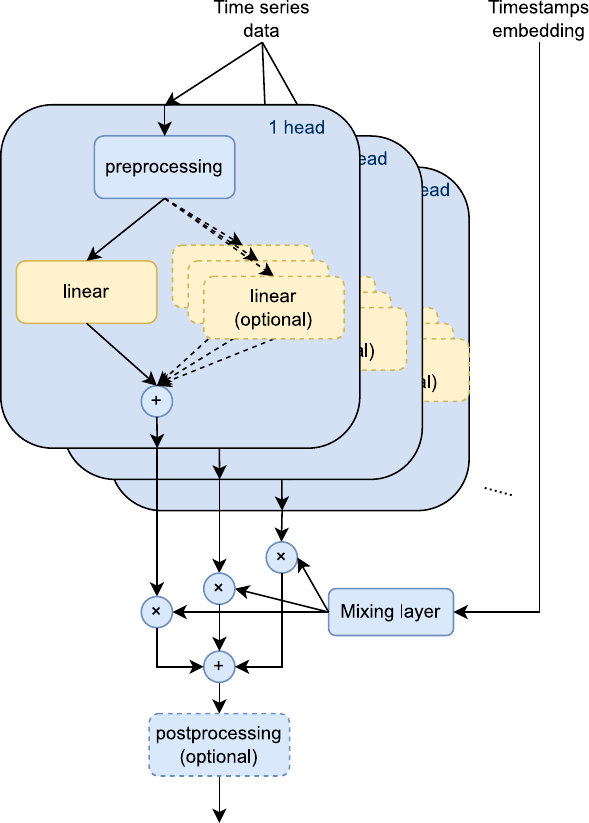}}
\vspace{.3in}
\caption{Structure of \nameshort{}. 
}
\label{fig:our_method}
\end{figure}

Our method, \nameshort{}, acts as a plugin for existing and potential future linear-centric LTSF models. A typical linear LTSF model can be represented as shown in Figure \ref{fig:common_linear}. Figure \ref{fig:our_method} illustrates how our mixture-of-experts method integrates with an arbitrary linear-centric LTSF model. Initially, we group everything up to and including the linear layers into one head. Input time series data is passed to all heads, and the outputs from all heads are passed to the mixing layer (which acts as the router in mixture-of-experts). The mixing layer comprises a two-layered MLP, which accepts the embedding of the starting timestamp as input and produces one weight for the outputs of each head. These weights are channel-specific, meaning each channel will have distinct weight sets that sum to 1.

More precisely, this method can be described as follows: Let $\mathcal{H}_i$ represent the i-th head, where $\mathcal{H}_i: \mathbb{R}^{c\times s} \rightarrow \mathbb{R}^{c\times p}$, $c$ is the number of channels, $s$ is the input sequence length, and $p$ is the prediction length. Let $\mathcal{M}$ be the MLP layer in the mixing layer, where $\mathcal{M}: \mathbb{R}^{t} \rightarrow \mathbb{R}^{c\times n}$, $t$ being the length of the embedding of the first timestamp, and $n$ is the number of heads. Let $\mathcal{P}$ be the postprocessing layer, where $\mathcal{P}: \mathbb{R}^{c\times p} \rightarrow \mathbb{R}^{c\times p}$. If $\mathbf{X} \in \mathbb{R}^{c\times s}$ is the input time series data and $\mathbf{X}_{mark} \in \mathbb{R}^{t}$ is the embedding of the first element of the input time series, the output ($\mathbf{Z}$) of the entire system can be expressed as:

\begin{eqnarray*}
\mathbf{Y}_i = \mathcal{H}_i(\mathbf{X});~~\mathbf{W} = \mathcal{M}(\mathbf{X}_{mark}) \\
\mathbf{Z} = \mathcal{P}(\sum_{i=1}^{n} \mathbf{W}_{:,i} \otimes \mathbf{Y}_i)
\end{eqnarray*}

Here, $\otimes$ is defined such that for a vector \( \mathbf{a} \in \mathbb{R}^c \) and a matrix \( \mathbf{B} \in \mathbb{R}^{c \times p} \), the operation \( \mathbf{a} \otimes \mathbf{B} \) yields a new matrix \( \mathbf{C} \in \mathbb{R}^{c \times p} \) where \( C_{ij} = a_i \cdot B_{ij} \) for all \( i \) and \( j \).

Compared with the conventional Mixture of Experts (MoE) architecture where experts' outputs are combined using a single weight, our proposed methodology (\nameshort{}) introduces channel-wise awareness by having the mixing layer output a unique set of weights for \emph{each} channel. In addition, instead of using the entire input data to determine gating, \nameshort{} only uses the embedding of the first timestamp for gating. This reduces the overall number of parameters and exploits the structure of long-term time series, which typically exhibit (possibly non-stationary) time-dependence.

In our experiments, we use a simple linear normalization methods to embed datetime values, where various temporal components of a datetime value are encoded into uniformly spaced values between $[-0.5, 0.5]$. We include the details of such embedding and the temporal components we chose for each dataset in Appendix \ref{app:embedding_time_stamps} (Supplementary Material).

\subsection{Toy Dataset}

\begin{figure}[h]
\vspace{.2in}
\centerline{\includeinkscape[width=0.4\textwidth]{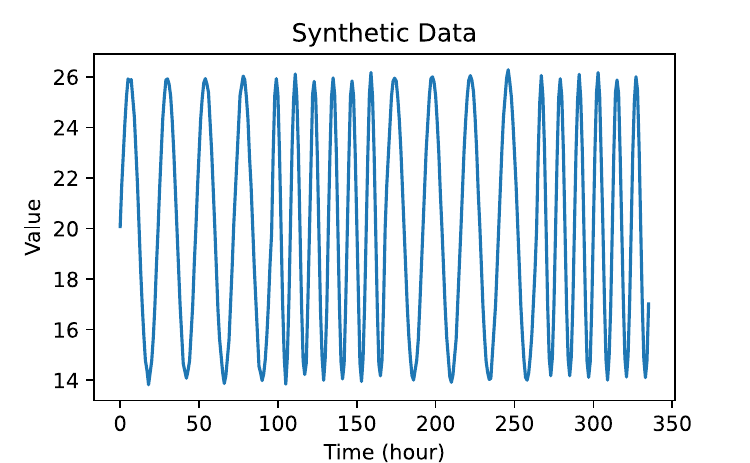}}
\vspace{.2in}
\caption{Visualization of the first 2 weeks (336 hours) of the toy dataset.}
\label{fig:toy_dataset}
\end{figure}

To show the intuitive effect of this approach, we first apply it to a toy dataset. 
We constructed the dataset such that from Monday to Thursday, the values adhere to a sinusoid function with frequency $f$. From Friday to Sunday, the values follow a sinusoid function with a doubled frequency of $2f$. Both have Gaussian noise added at each time step. This dataset exhibits weekly periodicity. The exact distribution is provided in \ref{toy_dataset_description} in Supplementary Materials, and the first two weeks of the datasets are visualized in Figure \ref{fig:toy_dataset}. 

\paragraph{Experimental Setup} For this experiment, we use RLinear as the backbone. The setup includes an input sequence length of 24 (1 day), a prediction length of 24 (1 day), a batch size of 128, and an initial learning rate of 0.005. We restrict ourselves to only 2 heads for the \nameshort{}-RLinear model in this experiment.

\begin{figure}[h!]
\vspace{0in}
\centerline{\includeinkscape[width=0.5\textwidth]{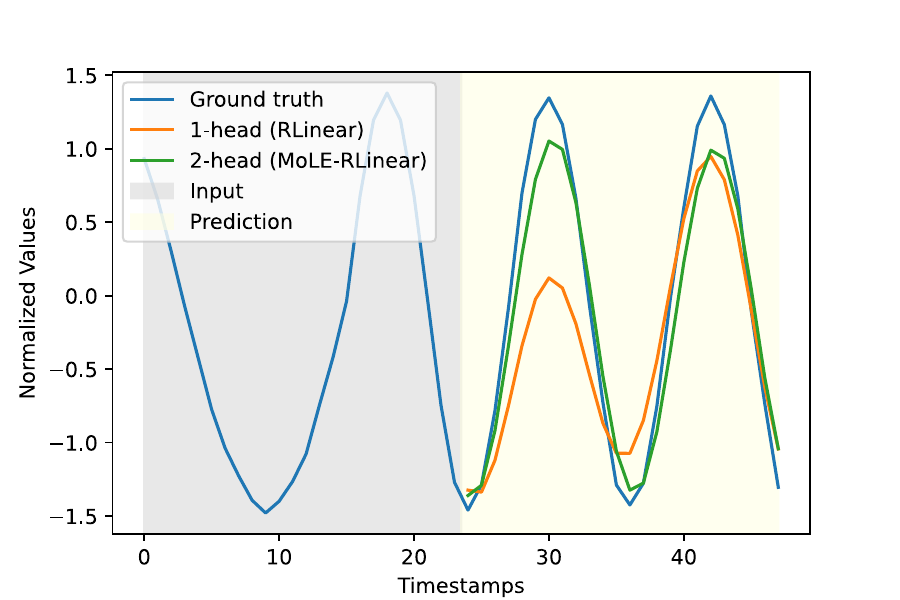}}
\vspace{.2in}
\caption{Predictions of single-head RLinear model and multi(2)-head \nameshort{}-RLinear model. The data from $t=0$ to $t=23$ is the input sequence, and the curves from $t=24$ onwards represent the ground truth and the predicted data from both methods. 
}
\label{fig:toy_dataset_output}
\end{figure}

\paragraph{Results} Figure \ref{fig:toy_dataset_output} demonstrates that the single-head RLinear model struggles to adapt to the pattern shift between Thursday and Friday.  Even though the input sequence carries information (within timestamps 20-23) indicating the transition to Friday, the single-head RLinear model, due to its single-layer design, fails to make accurate predictions. 
In contrast, the 2-head \nameshort{}-RLinear model captures this change well.

\section{EXPERIMENTS}

\subsection{Experimental Setup}

\paragraph{Datasets}
Following the standard practice of prior work \citep{wu2021autoformer,zhou2021informer,zeng2023transformers,li2023revisiting}, we conducted experiments on seven real-world datasets: the Electricity Transformer Temperature (ETT) datasets (ETTh1, ETTh2, ETTm1, ETTm2), as well as weather forecasting (Weather) \citep{angryk2020multivariate}, electricity forecasting (Electricity) \citep{khan2020towards}, and traffic flow estimation (Traffic)  \citep{chen2001freeway}. 
We also include a very recent time series dataset, Weather2K \citep{zhu2023weather2k}, which encompasses nearly three years of weather data from 1865 locations across China, featuring various weather conditions. We randomly selected four of these locations for evaluation: 79, 114, 850, 1786, which correspond to cities in the northeast, northwest, southwest, and south of China, respectively.

Table \ref{tab:datasets} in Supplementary Materials provides a brief overview of these datasets. For the ETT datasets, data is split into training, validation, and test sets in a temporal order with a ratio of 6:2:2. For all other datasets, the split ratio is 7:1:2. The split ratios we have employed in our experiments are consistent with the methodologies used in prior work.

\paragraph{Baseline Models}

We applied \nameshort{} to three recent linear-centric LTSF models: DLinear \citep{zeng2023transformers}, RLinear \citep{li2023revisiting}, and RMLP \citep{li2023revisiting}. We refer to the original models as \textit{original} and the models enhanced by our method as \textit{\nameshort{}}. Additionally, we evaluated the state-of-the-art transformer model, PatchTST \citep{nie2022time}. 
We evaluated all models on each dataset and reproduced results for linear-centric models. For PatchTST, we relied on author-reported outcomes. PatchTST's results for the newly released Weather2K dataset are not included, as it was not assessed in the original study.
In our comparisons, we used the results of supervised PatchTST/64 and PatchTST/42. PatchTST/64 means the number of input patches is 64, which uses the look-back window $L = 512$. PatchTST/42 means the number of input patches is 42, which has the default look-back window $L = 336$.
Our results on linear-centric models were mostly similar to reported results, so we use our computed values for our evaluation. 
In Appendix \ref{app:main_results} (Supplementary Material), we give a complete table that compares the reported results from prior work to the values we obtained experimentally, and provide additional discussion of these variations.
Additionally, we include the outcomes of further hyperparameter tuning to assess their impact on model performance in Appendix \ref{app:full_hp_tuning}.

\paragraph{Evaluation Metric}

As in existing studies, we adopt the widely-used evaluation metric, mean squared error (MSE), which has been utilized in the works of \cite{wu2021autoformer}, \cite{zhou2021informer}, \cite{zhang2022crossformer}, \cite{zeng2023transformers}, and \cite{li2023revisiting}. The metric is defined as:
$\operatorname{MSE} = \frac{1}{N}\sum_{i=1}^N (\mathbf{Y}_i - \hat{\mathbf{Y}}_i)^2$
where $N$ represents the total number of samples, $\mathbf{Y}_i$ indicates the i-th prediction, and $\hat{\mathbf{Y}}_i$ is the corresponding observation (ground truth).

\paragraph{Hyperparameters}

For all our experiments, including ablations, we employed a grid search approach to determine the optimal hyperparameters for each setting, as detailed in Table \ref{tab:hyperparameters}. For every experiment, we utilized the set of hyperparameters that produced the lowest validation loss to report the test loss.

\emph{Sequence and prediction length.}
As in prior work \cite{zeng2023transformers,li2023revisiting}, we fix the input sequence length to 336. We vary the prediction length, or the number of time stamps for which we predict data, in the set $\{96,192,336,720\}$. We  study the effect of variable input sequence length in Section \ref{sec:ablations}.

\begin{table}[h]
\caption{Hyperparameter search values.} \label{tab:hyperparameters}
\begin{center}
\begin{tabular}{ll}
\textbf{Hyperparameter}  &\textbf{Values} \\
\hline \\
Batch size         &8 \\
Initial learning rate             &0.005, 0.01, 0.05 \\
Number of heads             &  2, 3, 4, 5, 6\\
Head dropout rate & 0, 0.2 \\
Input sequence length & 336 \\
Prediction length &  \{96,192,336,720\}
\end{tabular}
\end{center}
\end{table}

\emph{Batch size.} To narrow down  our grid search, we conducted experiments to study how batch size influences model performance (Section \ref{sec:ablations}). Based on our findings, we  use a batch size of 8 for all experiments.

\emph{Initial learning rate.} The choice of initial learning rate also impacts model performance. For our experiments, we selected three rates: 0.005, 0.01, and 0.05.

\emph{Number of heads.} The number of heads in \nameshort{}  requires  tuning to cater to the distinct characteristics of datasets. We swept between 2-6 heads.

\emph{Head dropout.} To combat overfitting, we incorporated head dropout into \nameshort{}. Throughout training, the weights originating from the mixing layer are randomly dropped at a rate of $r$. The retained weights are adjusted to ensure their combined sum is 1. We tested two rates: 0 (head dropout deactivated) and 0.2. We also provide a detailed analysis and comprehensive results on how the choice of dropout rate impacts multi-head model performance in Appendix \ref{app:dropout_rate}.

\paragraph{Implementation Details}
For the conducted experiments, the mixing layer comprises a 2-layer MLP. The hidden layer dimension matches that of the output layer. A \textit{ReLU} activation lies between the layers.

\subsection{Comparison with Single-Head Models}

\begin{table*}[h!]
\caption{Comparison between original (single-head) and enhanced (multi-head \nameshort{}) DLinear, RLinear, and RMLP models. Prediction length $\in \{96, 192, 336, 720\}$. The values reported are MSE loss. A lower value indicates a better prediction. The better results among a linear-centric method and its \nameshort{} variant are highlighted in \textbf{bold.} The cells shown in \textcolor{blue}{blue} indicate the current SOTA for a given dataset and prediction length. 
For all three linear-centric models, \nameshort{} improves forecasting error in over 78\% of the datasets and settings we evaluated.
Entries marked with `--' indicate datasets and prediction lengths not reported in previous work.
} \label{tab:dlinear}
\begin{center}
\begin{adjustbox}{width=1.76\columnwidth,center}
\begin{tabular}{lc|cc|cc|cc|}
\toprule
\multicolumn{2}{c}{\textbf{Model}} &\multicolumn{2}{c}{DLinear} &\multicolumn{2}{c}{RLinear} &\multicolumn{2}{c}{RMLP} \\\cmidrule{1-8}
\multicolumn{2}{c}{Dataset ~~ Prediction length} &original &\nameshort{} &original &\nameshort{} &original &\nameshort{} \\\midrule

\multirow{4}{*}{ETTh1} &96 &\textbf{0.372} &0.377 &\textbf{0.371} &0.375 &\textbf{0.381} &0.421\\
&192 &\textbf{0.413} &0.453 &0.404 &\color{blue}\textbf{0.403} &0.541 &\textbf{0.432}\\
&336 &\textbf{0.442} &0.469 &\textbf{0.428} &0.430 &0.453 &\textbf{0.437}\\
&720 &\textbf{0.501} &0.505 &0.450 &\textbf{0.449} &0.502 &\textbf{0.474}\\
\midrule
\multirow{4}{*}{ETTh2} &96 &{\textbf{0.287}} &\textbf{0.287} &\color{blue}\textbf{0.272} &0.273 &\textbf{0.294} &0.322 \\
&192 &\textbf{0.349} &0.362 &0.341 &\color{blue}\textbf{0.336} &\textbf{0.362} &0.375 \\
&336 &0.430 &\textbf{0.419} &0.372 &\textbf{0.371} &\textbf{0.389} &0.416 \\
&720 &0.710 &\textbf{0.605} &0.418 &\textbf{0.409} &0.440 &\textbf{0.418} \\
\midrule
\multirow{4}{*}{ETTm1} &96 &0.300 &\color{blue}\textbf{0.286} &0.301 &\textbf{0.291} &0.300 &\textbf{0.294} \\
&192 &0.336 &\color{blue}\textbf{0.328} &0.335 &\textbf{0.333} &\textbf{0.339} &0.340 \\
&336 &\textbf{0.374} &0.380 &0.371 &\textbf{0.368} &{\color{blue}\textbf{0.365}}&\color{blue}\textbf{0.365}\\
&720 &0.461 &\textbf{0.447} &\textbf{0.429} &{\textbf{0.429}} &0.439 &\textbf{0.426}\\
\midrule
\multirow{4}{*}{ETTm2} &96 &{\textbf{0.168}} &\textbf{0.168} &0.164 &\color{blue}\textbf{0.163} &0.165 &\textbf{0.163} \\
&192 &\textbf{0.228} &0.233 &0.219 &\color{blue}\textbf{0.217} &0.223 &\textbf{0.220} \\
&336 &0.295 &\textbf{0.289} &{\color{blue}\textbf{0.272}} &\color{blue}\textbf{0.272} &\textbf{0.282} &{\textbf{0.282}}\\
&720 &\textbf{0.382} &0.399 &\textbf{0.368} &0.380 &\color{blue}\textbf{0.362} &0.371\\
\midrule
\multirow{4}{*}{weather} &96 &0.175 &\color{blue}\textbf{0.147} &0.174 &\textbf{0.152} &0.156 &\textbf{0.153} \\
&192 &0.224 &\textbf{0.203} &0.217 &\textbf{0.190} &0.203 &\color{blue}\textbf{0.190}\\
&336 &0.263 &\color{blue}\textbf{0.238} &0.264 &\textbf{0.245} &0.254 &\textbf{0.242} \\
&720 &0.332 &\color{blue}\textbf{0.314} &0.331 &\textbf{0.316} &0.331 &\textbf{0.323}\\
\midrule
\multirow{4}{*}{electricity} &96 &0.140 &\textbf{0.131} &0.143 &\textbf{0.133} &0.131 &\color{blue}\textbf{0.129} \\
&192 &0.153 &\color{blue}\textbf{0.147} &0.157 &\textbf{0.150} &\textbf{0.149} &0.152 \\
&336 &0.169 &\color{blue}\textbf{0.162} &0.174 &\textbf{0.164} &0.167 &\textbf{0.166} \\
&720 &0.203 &\textbf{0.180} &0.212 &\textbf{0.182} &0.200 &\color{blue}\textbf{0.178}\\
\midrule
\multirow{4}{*}{traffic} &96 &0.410 &\textbf{0.390} &0.412 &\textbf{0.384} &0.380 &\textbf{0.372} \\
&192 &0.423 &\textbf{0.397} &0.424 &\textbf{0.397} &0.396 &\textbf{0.385} \\
&336 &0.436 &\textbf{0.425} &0.437 &\textbf{0.415} &0.409 &\textbf{0.407} \\
&720 &0.466 &\textbf{0.446} &0.466 &\textbf{0.440} &0.441 &\color{blue}\textbf{0.429}\\
\midrule
\multirow{4}{*}{Weather2K79} &96 &0.571 &\color{blue}\textbf{0.555} &0.572 &\textbf{0.564} &0.584 &\textbf{0.567} \\
&192 &0.593 &\color{blue}\textbf{0.566} &0.595 &\textbf{0.588} &0.601 &\textbf{0.588} \\
&336 &0.590 &\color{blue}\textbf{0.546} &0.592 &\textbf{0.575} &0.594 &\textbf{0.574} \\
&720 &0.619 &\color{blue}\textbf{0.535} &0.624 &\textbf{0.566} &0.616 &\textbf{0.565} \\
\midrule
\multirow{4}{*}{Weather2K114} &96 &0.409 &\color{blue}\textbf{0.391} &0.407 &\textbf{0.395} &\textbf{0.403} &\textbf{0.403} \\
&192 &0.437 &\color{blue}\textbf{0.405} &0.436 &\textbf{0.427} &0.438 &\textbf{0.424} \\
&336 &0.460 &\color{blue}\textbf{0.415} &0.459 &\textbf{0.439} &0.453 &\textbf{0.441} \\
&720 &0.506 &\color{blue}\textbf{0.425} &0.509 &\textbf{0.482} &0.495 &\textbf{0.476} \\
\midrule
\multirow{4}{*}{Weather2K850} &96 &0.481 &\textbf{0.474} &0.483 &\color{blue}\textbf{0.471} &\textbf{0.481} &0.483 \\
&192 &0.502 &\color{blue}\textbf{0.484} &0.505 &\textbf{0.495} &0.509 &\textbf{0.495} \\
&336 &0.509 &\color{blue}\textbf{0.474} &0.513 &\textbf{0.502} &0.513 &\textbf{0.499} \\
&720 &0.523 &\color{blue}\textbf{0.461} &0.527 &\textbf{0.489} &0.527 &\textbf{0.491} \\
\midrule
\multirow{4}{*}{Weather2K1786} &96 &0.545 &\color{blue}\textbf{0.535} &0.545 &\color{blue}\textbf{0.535} &0.550 &\textbf{0.544} \\
&192 &\textbf{0.591} &0.601 &0.591 &\color{blue}\textbf{0.581} &0.600 &\textbf{0.584} \\
&336 &0.620 &\color{blue}\textbf{0.603} &0.620 &\textbf{0.618} &0.634 &\textbf{0.617} \\
&720 &\textbf{0.658} &0.660 &0.660 &\color{blue}\textbf{0.628} &0.668 &\textbf{0.640} \\
\midrule

\multicolumn{2}{c}{No. improved (\%)} &\multicolumn{2}{c}{32 (72.7\%)} &\multicolumn{2}{c}{38 (86.4\%)} &\multicolumn{2}{c}{33 (75.0\%)}\\

\bottomrule
\end{tabular}%
\begin{tabular}{cc}
\toprule
\multicolumn{2}{c}{PatchTST} \\\cmidrule{1-2}
64 &42\\\midrule
\color{blue}0.370& 0.375\\
0.413& 0.414\\
\color{blue}0.422& 0.431\\
\color{blue}0.447& 0.449\\\midrule
0.274& 0.274\\
0.341& 0.339\\
\color{blue}0.329& 0.331\\
\color{blue}0.379& \color{blue}0.379\\\midrule
0.293& 0.290\\
0.333& 0.332\\
0.369& 0.366\\
\color{blue}0.416& 0.420\\\midrule
0.166& 0.165\\
0.223& 0.220\\
0.274& 0.278\\
\color{blue}0.362& 0.367\\\midrule
0.149& 0.152\\
0.194& 0.197\\
0.245& 0.249\\
\color{blue}0.314& 0.320\\\midrule
\color{blue}0.129& 0.130\\
\color{blue}0.147& 0.148\\
0.163& 0.167\\
0.197& 0.202\\\midrule
\color{blue}0.360& 0.367\\
\color{blue}0.379& 0.385\\
\color{blue}0.392& 0.398\\
0.432& 0.434\\\midrule
--&--\\--&--\\--&--\\--&--\\\midrule
--&--\\--&--\\--&--\\--&--\\\midrule
--&--\\--&--\\--&--\\--&--\\\midrule
--&--\\--&--\\--&--\\--&--\\\midrule\\
\bottomrule
\end{tabular}
\end{adjustbox}
\end{center}
\end{table*}

Table \ref{tab:dlinear} presents the evaluation results for three leading linear-centric models: DLinear, RLinear, and RMLP,  across various datasets and prediction length settings. 
{Among the 44 dataset and prediction length combinations, \nameshort{} improves  the forecasting error of the original model in \textbf{32/44 (73\% of experiments)} instances for DLinear, \textbf{38/44 (86\%)} instances for RLinear, and \textbf{33/44 (75\% of experiments)} instances for RMLP.}
In particular, we note that for larger datasets, such as weather, electricity, and traffic, \nameshort{} enhances all 12 (\textbf{100\%}) combinations of dataset and prediction length for DLinear and RLinear. The cells shown in \textcolor{blue}{blue} indicate the current SOTA for a given dataset and prediction length. Enhanced by \nameshort{}, these linear-centric models achieve SOTA in \textbf{68\%} of experimental settings for which PatchTST also reports numbers (to which we also compare), whereas the existing linear-centric models were previously SOTA in only {25}\% of cases. For the Weather2K datasets, \nameshort{}-linear-centric models achieve SOTA in \textbf{all} settings.

\subsection{Ablations and Design Choices}
\label{sec:ablations}

\paragraph{Does the timestamp input help the mixing layer learn better?}

To demonstrate that the performance enhancement from \nameshort{} is not just a result of increased model size, but rather the introduction of timestamp information, we designed the following  experiments: 
1) We replaced the timestamp embedding with random numbers (while keeping the values within the original range of the timestamp embedding). We call this variant \textit{RandomIn}.
2) Building on the first ablation, we introduced a second modification where we use random weights instead of the outputs from the mixing layer to compute a weighted sum of the outputs from the heads. This is done both at training and test time.
We call this variant \textit{RandomOut}. We refer to the original \nameshort{} with time stamp input as \textit{TimeIn}. 
For these design variants, we maintained the same hyperparameter search strategy shown in Table \ref{tab:hyperparameters}. Consistent with our previous experiments, we chose the set of hyperparameters that produced the best validation loss to compute the reported test loss. We ran experiments using the DLinear model.

\begin{table}[h!]
\caption{Comparison between \textit{TimeIn} (proposed method, conditioning on timestamp) with conditioning on a random input \textit{RandomIn} and randomly mixing outputs \textit{RandomOut}. 
The best MSE results are highlighted in \textbf{bold}, 2nd lowest losses are underlined. W2K stands for Weather2K datasets.} \label{tab:ablation}
\begin{center}

\begin{adjustbox}{width=0.8\columnwidth,center}

\begin{tabular}{lrrrrr}\toprule
\multicolumn{2}{c}{Data ~~ \begin{tabular}[c]{@{}l@{}}Pred.\\ Len.\end{tabular}} &TimeIn &RandomIn &RandomOut \\\midrule
\multirow{4}{*}{\rotatebox[origin=c]{90}{ETTh1}} &96 &0.3768 &\textbf{0.3750} &\ul{0.3755} \\
&192 &0.4531 &\ul{0.4526} &\textbf{0.4297} \\
&336 &\textbf{0.4689} &0.4742 &\ul{0.4740} \\
&720 &\textbf{0.5046} &0.5302 &\ul{0.5229} \\\midrule
\multirow{4}{*}{\rotatebox[origin=c]{90}{ETTh2}} &96 &\ul{0.2865} &0.2916 &\textbf{0.2859} \\
&192 &\ul{0.3617} &\textbf{0.3481} &0.3618 \\
&336 &0.4187 &\textbf{0.3864} &\ul{0.4156} \\
&720 &0.6053 &\textbf{0.5929} &\ul{0.6029} \\\midrule
\multirow{4}{*}{\rotatebox[origin=c]{90}{ETTm1}} &96 &\textbf{0.2862} &\ul{0.2865} &0.3031 \\
&192 &\textbf{0.3281} &\ul{0.3288} &0.3357 \\
&336 &0.3797 &\textbf{0.3712} &\ul{0.3750} \\
&720 &\ul{0.4466} &\textbf{0.4463} &0.4506 \\\midrule
\multirow{4}{*}{\rotatebox[origin=c]{90}{ETTm2}}  &96 &\ul{0.1677} &0.1678 &\textbf{0.1666} \\
&192 &0.2334 &\ul{0.2308} &\textbf{0.2252} \\
&336 &0.2889 &\ul{0.2828} &\textbf{0.2776} \\
&720 &\ul{0.3985} &0.4128 &\textbf{0.3807} \\\midrule
\multirow{4}{*}{\rotatebox[origin=c]{90}{Weather}}&96 &\ul{0.1466} &\textbf{0.1454} &0.1768 \\
&192 &\ul{0.2025} &\textbf{0.1872} &0.2240 \\
&336 &\textbf{0.2381} &\ul{0.2461} &0.2658 \\
&720 &\textbf{0.3142} &\ul{0.3249} &0.3335 \\\midrule
\multirow{4}{*}{\rotatebox[origin=c]{90}{Electricity}} &96 &\textbf{0.1314} &\ul{0.1332} &0.1399 \\
&192 &\ul{0.1474} &\textbf{0.1472} &0.1532 \\
&336 &\textbf{0.1618} &\ul{0.1625} &0.1688 \\
&720 &\textbf{0.1796} &\ul{0.1974} &0.2031 \\\midrule
\multirow{4}{*}{\rotatebox[origin=c]{90}{Traffic}} &96 &\textbf{0.3903} &\ul{0.4096} &0.4100 \\
&192 &\textbf{0.3966} &\ul{0.4221} &0.4226 \\
&336 &\textbf{0.4251} &\ul{0.4338} &0.4354 \\
&720 &\textbf{0.4460} &\ul{0.4631} &0.4655 \\\midrule
\multirow{4}{*}{\rotatebox[origin=c]{90}{W2K79}} &96 &\textbf{0.5547} &\ul{0.5565} &0.5711 \\
&192 &\textbf{0.5662} &\ul{0.5810} &0.5927 \\
&336 &\textbf{0.5465} &\ul{0.5779} &0.5900 \\
&720 &\textbf{0.5348} &\ul{0.6012} &0.6198 \\\midrule
\multirow{4}{*}{\rotatebox[origin=c]{90}{W2K114}} &96 &\ul{0.3913} &\textbf{0.3894} &0.4087 \\
&192 &\textbf{0.4053} &\ul{0.4204} &0.4371 \\
&336 &\textbf{0.4148} &\ul{0.4424} &0.4601 \\
&720 &\textbf{0.4246} &\ul{0.4907} &0.5066 \\\midrule
\multirow{4}{*}{\rotatebox[origin=c]{90}{W2K850}} &96 &\ul{0.4741} &\textbf{0.4648} &0.4804 \\
&192 &\textbf{0.4841} &\ul{0.4931} &0.5017 \\
&336 &\textbf{0.4745} &\ul{0.5018} &0.5091 \\
&720 &\textbf{0.4613} &\ul{0.5089} &0.5234 \\\midrule
\multirow{4}{*}{\rotatebox[origin=c]{90}{W2K1786}} &96 &\textbf{0.5346} &0.5564 &\ul{0.5459} \\
&192 &\ul{0.6014} &0.6025 &\textbf{0.5900} \\
&336 &\textbf{0.6034} &0.6345 &\ul{0.6177} \\
&720 &\ul{0.6603} &0.6827 &\textbf{0.6583} \\\midrule
\multicolumn{2}{c}{No. lowest} &25/44 &11/44 &8/44 \\
\bottomrule
\end{tabular}
\end{adjustbox}
\end{center}
\end{table}

Surprisingly, \textit{RandomIn} and \textit{RandomOut} occasionally produce losses lower than \textit{TimeIn}, which uses timestamp embeddings. 
This is consistent with the findings of \citet{roller2021hash} on MoE with hash layers.
We explore why in the next subsection. 
For now, we comment on the robustness of \textit{TimeIn}. As shown in Table \ref{tab:ablation}, across all experiments, \textit{TimeIn}  had the lowest loss in 25 instances, surpassing \emph{RandomIn} and \emph{RandomOut}, which achieved this 11 and 8 times, respectively.

\paragraph{When, and why does randomness help?}
We next try to understand when and why random inputs and outputs can help linear-centric models. 
We propose two possible effects contributing to this observation: \textbf{Effect 1}: \emph{RandomIn} and \emph{RandomOut} may have a regularization effect, akin to dropout \citep{baldi2013understanding}, which prevents the network from overfitting head weights to the training data. This has the effect of making \emph{all} heads better. 
\textbf{Effect 2}: Suppose time series can be roughly divided into regimes with different temporal patterns (e.g., weekdays vs weekends). For long input sequences, the sequence may traverse multiple regimes. Hence, no single head will be able to capture the full dynamics, regardless of start time. 
However, as the input sequence lengthens, the linear layer has more past data to consider, and some patterns in this data can hint at upcoming changes in the series. 
Hence, we predict that the effect of conditioning on start time should be less beneficial. Conversely, as the input sequence length is reduced, the effect of input time should be more beneficial.

We hypothesize that both effects contribute to the observations  in Table \ref{tab:ablation}.
To validate this hypothesis, we conducted the following experiment on \nameshort{-DLinear}. 
To test effect 1, we compared \emph{TimeIn}, \emph{RandomIn}, and \emph{RandomOut} to a variant in which we conduct dropout \emph{over the heads}. That is, during training, we  exclude in each iteration a randomly-selected fraction (20\%)  of heads. At test time, we use all heads according to the learned weights in the mixing layer.
To test effect 2, we fixed the prediction length to 100 and progressively varied the input sequence length among $\{6, 88, 170, 254, 336\}$. 

Following the same methodology as our main experiments, we performed a grid search on hyperparameters and reported the test loss of the best set of hyperparameters for each seed. 
Each experiment is averaged over three runs over the Electricity dataset.

\begin{figure}[t!]
\vspace{0.1in}
\centerline{\includeinkscape[width=0.47\textwidth]{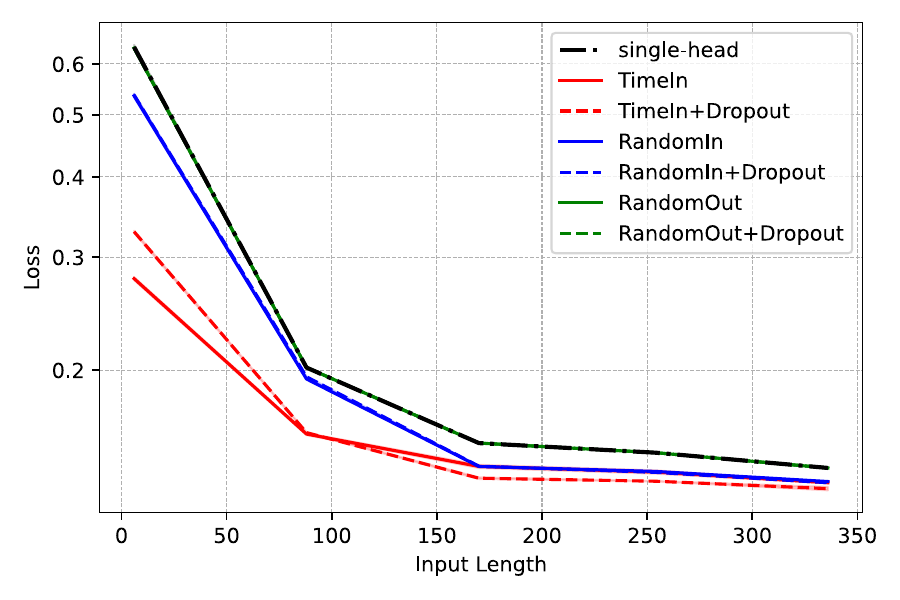}}
\vspace{.3in}
\caption{Performance comparison of various methods based on input length on \nameshort{-DLinear}.}
\label{fig:small_input}
\end{figure}

In Figure \ref{fig:small_input}, we plot the mean MSE for each input length, and the shadows around the curves represent the $\pm$ standard deviation. The shadows are not clearly visible since the results are stable.
Figure \ref{fig:small_input} illustrates three important trends: 

\noindent (1) As input length increases, the performance of \emph{TimeIn} approaches that of the single-head baseline, as well as \emph{RandomIn} and \emph{RandomOut}. 
This suggests that Effect 2 may indeed be correct: conditioning on timestamp is more beneficial as input sequence length decreases. (Note that our earlier results in Table \ref{tab:dlinear} used a fixed input length of 336, which is the rightmost point of Figure \ref{fig:small_input}, where the gains of \nameshort{} are least pronounced.)

\begin{figure}[t!]
\vspace{.1in}
\centerline{\includeinkscape[width=0.5\textwidth]{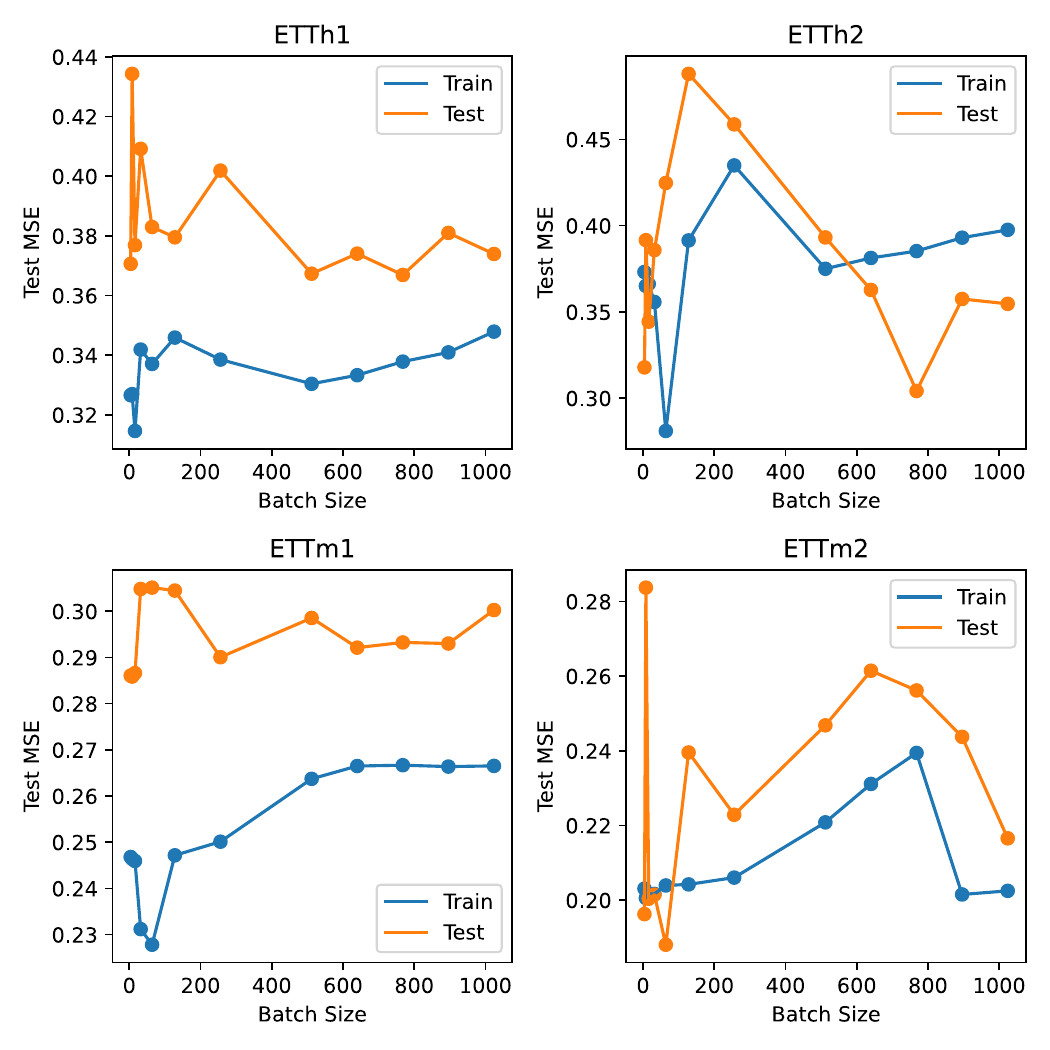}}
\vspace{.3in}
\caption{Impact of batch size on \nameshort{-DLinear} 
}
\label{fig:batch_size}
\end{figure}

\noindent (2) Dropout has virtually no effect on \emph{RandomIn} or \emph{RandomOut}, but it does affect \emph{TimeIn}. This suggests that using random inputs may indeed have a regularization effect (Effect 1), which is roughly equivalent to dropout over heads. To take advantage of both effects at once, we can condition on time stamp while also doing dropout over heads.

\noindent (3) As input sequence length grows, the effect of dropout on \emph{TimeIn} becomes more beneficial. This suggests that overfitting to the start timestamp is more detrimental for longer input sequences. 

\paragraph{How does batch size influence model performance and what size should we use?}

We finally present the results of our experiments to determine the batch size for our prior experiments. Figure \ref{fig:batch_size} plots how the final test loss varies with batch size, for four datasets. For these experiments, we used DLinear as the base model, with an input sequence length of 336, prediction length of 96, five heads, and an initial learning rate of 0.05.
We observe that in general, test losses are lower when batch sizes are either very small or (in some cases) large (relative to intermediate sizes ranging from 16 to 128). 
This mid-range of batch sizes exhibits poor generalization, with low training loss but high test loss, similar to observations in  \cite{keskar2016large}. For robustness, 
we used a batch size of 8.
Comprehensive results and a more detailed analysis of batch size effects are available in Appendix \ref{app:batch_size}.

\section{CONCLUSION AND FUTURE WORK}

We propose \nameshort{}, a technique based on mixture-of-experts that can be applied out-of-the-box to existing linear-centric models to improve performance in the LTSF problem. 
Our results strongly suggest that \nameshort{} gives significant and robust gains across a wide range of datasets and prediction settings. 
An important open question is precisely characterizing high-dimensional datasets to understand exactly when \nameshort{} will help. 
Although we hypothesize that this question relates to 
% the degree and patterns of nonstationarity 
the variability and complexity of patterns
in the time series dataset, it is currently unclear how to predict how many heads are needed.

\section*{Acknowledgments}
The authors gratefully acknowledge the support of the National Science Foundation (RINGS program) under grant CNS-2148359. This work was also made possible by the generous support of C3.ai, Bosch, Cisco, Intel, J.P. Morgan Chase, the Sloan Foundation, and Siemens.  

\bibliography{mybibtex}

\begin{thebibliography}{48}
\providecommand{\natexlab}[1]{#1}
\providecommand{\url}[1]{\texttt{#1}}
\expandafter\ifx\csname urlstyle\endcsname\relax
  \providecommand{\doi}[1]{doi: #1}\else
  \providecommand{\doi}{doi: \begingroup \urlstyle{rm}\Url}\fi

\bibitem[Angryk et~al.(2020)Angryk, Martens, Aydin, Kempton, Mahajan, Basodi, Ahmadzadeh, Cai, Filali~Boubrahimi, Hamdi, et~al.]{angryk2020multivariate}
Rafal~A Angryk, Petrus~C Martens, Berkay Aydin, Dustin Kempton, Sushant~S Mahajan, Sunitha Basodi, Azim Ahmadzadeh, Xumin Cai, Soukaina Filali~Boubrahimi, Shah~Muhammad Hamdi, et~al.
\newblock Multivariate time series dataset for space weather data analytics.
\newblock \emph{Scientific data}, 7\penalty0 (1):\penalty0 227, 2020.

\bibitem[Ariyo et~al.(2014)Ariyo, Adewumi, and Ayo]{ariyo2014stock}
Adebiyi~A Ariyo, Adewumi~O Adewumi, and Charles~K Ayo.
\newblock Stock price prediction using the arima model.
\newblock In \emph{2014 UKSim-AMSS 16th international conference on computer modelling and simulation}, pages 106--112. IEEE, 2014.

\bibitem[Artetxe et~al.(2021)Artetxe, Bhosale, Goyal, Mihaylov, Ott, Shleifer, Lin, Du, Iyer, Pasunuru, et~al.]{artetxe2021efficient}
Mikel Artetxe, Shruti Bhosale, Naman Goyal, Todor Mihaylov, Myle Ott, Sam Shleifer, Xi~Victoria Lin, Jingfei Du, Srinivasan Iyer, Ramakanth Pasunuru, et~al.
\newblock Efficient large scale language modeling with mixtures of experts.
\newblock \emph{arXiv preprint arXiv:2112.10684}, 2021.

\bibitem[Bai et~al.(2018)Bai, Kolter, and Koltun]{bai2018empirical}
Shaojie Bai, J~Zico Kolter, and Vladlen Koltun.
\newblock An empirical evaluation of generic convolutional and recurrent networks for sequence modeling.
\newblock \emph{arXiv preprint arXiv:1803.01271}, 2018.

\bibitem[Baldi and Sadowski(2013)]{baldi2013understanding}
Pierre Baldi and Peter~J Sadowski.
\newblock Understanding dropout.
\newblock \emph{Advances in neural information processing systems}, 26, 2013.

\bibitem[Borovykh et~al.(2017)Borovykh, Bohte, and Oosterlee]{borovykh2017conditional}
Anastasia Borovykh, Sander Bohte, and Cornelis~W Oosterlee.
\newblock Conditional time series forecasting with convolutional neural networks.
\newblock \emph{arXiv preprint arXiv:1703.04691}, 2017.

\bibitem[BOX(1976)]{box1976time}
G~BOX.
\newblock Time series analysis forecasting and control.
\newblock \emph{Holden-Day, Oackland. California 1976}, 1976.

\bibitem[Chang et~al.(2018)Chang, Sun, Wu, and Lin]{chang2018memory}
Yen-Yu Chang, Fan-Yun Sun, Yueh-Hua Wu, and Shou-De Lin.
\newblock A memory-network based solution for multivariate time-series forecasting.
\newblock \emph{arXiv preprint arXiv:1809.02105}, 2018.

\bibitem[Chen et~al.(2001)Chen, Petty, Skabardonis, Varaiya, and Jia]{chen2001freeway}
Chao Chen, Karl Petty, Alexander Skabardonis, Pravin Varaiya, and Zhanfeng Jia.
\newblock Freeway performance measurement system: mining loop detector data.
\newblock \emph{Transportation Research Record}, 1748\penalty0 (1):\penalty0 96--102, 2001.

\bibitem[Chen et~al.(2023)Chen, Li, Yoder, Arik, and Pfister]{chen2023tsmixer}
Si-An Chen, Chun-Liang Li, Nate Yoder, Sercan~O Arik, and Tomas Pfister.
\newblock Tsmixer: An all-mlp architecture for time series forecasting.
\newblock \emph{arXiv preprint arXiv:2303.06053}, 2023.

\bibitem[Das et~al.(2023)Das, Kong, Leach, Sen, and Yu]{das2023long}
Abhimanyu Das, Weihao Kong, Andrew Leach, Rajat Sen, and Rose Yu.
\newblock Long-term forecasting with tide: Time-series dense encoder.
\newblock \emph{arXiv preprint arXiv:2304.08424}, 2023.

\bibitem[Dryden and Hoefler(2022)]{dryden2022spatial}
Nikoli Dryden and Torsten Hoefler.
\newblock Spatial mixture-of-experts.
\newblock \emph{Advances in Neural Information Processing Systems}, 35:\penalty0 11697--11713, 2022.

\bibitem[Du et~al.(2023)Du, Su, and Wei]{du2023preformer}
Dazhao Du, Bing Su, and Zhewei Wei.
\newblock Preformer: predictive transformer with multi-scale segment-wise correlations for long-term time series forecasting.
\newblock In \emph{ICASSP 2023-2023 IEEE International Conference on Acoustics, Speech and Signal Processing (ICASSP)}, pages 1--5. IEEE, 2023.

\bibitem[Fan et~al.(2019)Fan, Zhang, Pan, Li, Zhang, Yuan, Wu, Wang, Pei, and Huang]{fan2019multi}
Chenyou Fan, Yuze Zhang, Yi~Pan, Xiaoyue Li, Chi Zhang, Rong Yuan, Di~Wu, Wensheng Wang, Jian Pei, and Heng Huang.
\newblock Multi-horizon time series forecasting with temporal attention learning.
\newblock In \emph{Proceedings of the 25th ACM SIGKDD International conference on knowledge discovery \& data mining}, pages 2527--2535, 2019.

\bibitem[Fildes(1991)]{fildes1991forecasting}
Robert Fildes.
\newblock Forecasting, structural time series models and the kalman filter: Bayesian forecasting and dynamic models.
\newblock \emph{Journal of the Operational Research Society}, 42\penalty0 (11):\penalty0 1031--1033, 1991.

\bibitem[Gong et~al.(2023)Gong, Tang, and Liang]{gong2023patchmixer}
Zeying Gong, Yujin Tang, and Junwei Liang.
\newblock Patchmixer: A patch-mixing architecture for long-term time series forecasting.
\newblock \emph{arXiv preprint arXiv:2310.00655}, 2023.

\bibitem[Gruver et~al.(2023)Gruver, Finzi, Qiu, and Wilson]{gruver2023large}
Nate Gruver, Marc Finzi, Shikai Qiu, and Andrew~Gordon Wilson.
\newblock Large language models are zero-shot time series forecasters.
\newblock \emph{arXiv preprint arXiv:2310.07820}, 2023.

\bibitem[Han et~al.(2019)Han, Zhao, Leung, Ma, and Wang]{han2019review}
Zhongyang Han, Jun Zhao, Henry Leung, King~Fai Ma, and Wei Wang.
\newblock A review of deep learning models for time series prediction.
\newblock \emph{IEEE Sensors Journal}, 21\penalty0 (6):\penalty0 7833--7848, 2019.

\bibitem[Jiang et~al.(2023)Jiang, Han, Zhao, and Wang]{pdformer}
Jiawei Jiang, Chengkai Han, Wayne~Xin Zhao, and Jingyuan Wang.
\newblock Pdformer: Propagation delay-aware dynamic long-range transformer for traffic flow prediction.
\newblock In \emph{{AAAI}}. {AAAI} Press, 2023.

\bibitem[Jordan and Jacobs(1994)]{jordan1994hierarchical}
Michael~I Jordan and Robert~A Jacobs.
\newblock Hierarchical mixtures of experts and the em algorithm.
\newblock \emph{Neural computation}, 6\penalty0 (2):\penalty0 181--214, 1994.

\bibitem[Keskar et~al.(2016)Keskar, Mudigere, Nocedal, Smelyanskiy, and Tang]{keskar2016large}
Nitish~Shirish Keskar, Dheevatsa Mudigere, Jorge Nocedal, Mikhail Smelyanskiy, and Ping Tak~Peter Tang.
\newblock On large-batch training for deep learning: Generalization gap and sharp minima.
\newblock \emph{arXiv preprint arXiv:1609.04836}, 2016.

\bibitem[Khan et~al.(2020)Khan, Hussain, Ullah, Rho, Lee, and Baik]{khan2020towards}
Zulfiqar~Ahmad Khan, Tanveer Hussain, Amin Ullah, Seungmin Rho, Miyoung Lee, and Sung~Wook Baik.
\newblock Towards efficient electricity forecasting in residential and commercial buildings: A novel hybrid cnn with a lstm-ae based framework.
\newblock \emph{Sensors}, 20\penalty0 (5):\penalty0 1399, 2020.

\bibitem[Kim et~al.(2021)Kim, Kim, Tae, Park, Choi, and Choo]{kim2021reversible}
Taesung Kim, Jinhee Kim, Yunwon Tae, Cheonbok Park, Jang-Ho Choi, and Jaegul Choo.
\newblock Reversible instance normalization for accurate time-series forecasting against distribution shift.
\newblock In \emph{International Conference on Learning Representations}, 2021.

\bibitem[Lai et~al.(2018)Lai, Chang, Yang, and Liu]{lai2018modeling}
Guokun Lai, Wei-Cheng Chang, Yiming Yang, and Hanxiao Liu.
\newblock Modeling long-and short-term temporal patterns with deep neural networks.
\newblock In \emph{The 41st international ACM SIGIR conference on research \& development in information retrieval}, pages 95--104, 2018.

\bibitem[Li et~al.(2023)Li, Qi, Li, and Xu]{li2023revisiting}
Zhe Li, Shiyi Qi, Yiduo Li, and Zenglin Xu.
\newblock Revisiting long-term time series forecasting: An investigation on linear mapping.
\newblock \emph{arXiv preprint arXiv:2305.10721}, 2023.

\bibitem[Liu et~al.(2021)Liu, Yu, Liao, Li, Lin, Liu, and Dustdar]{liu2021pyraformer}
Shizhan Liu, Hang Yu, Cong Liao, Jianguo Li, Weiyao Lin, Alex~X Liu, and Schahram Dustdar.
\newblock Pyraformer: Low-complexity pyramidal attention for long-range time series modeling and forecasting.
\newblock In \emph{International conference on learning representations}, 2021.

\bibitem[Nie et~al.(2022)Nie, Nguyen, Sinthong, and Kalagnanam]{nie2022time}
Yuqi Nie, Nam~H Nguyen, Phanwadee Sinthong, and Jayant Kalagnanam.
\newblock A time series is worth 64 words: Long-term forecasting with transformers.
\newblock \emph{arXiv preprint arXiv:2211.14730}, 2022.

\bibitem[Riquelme et~al.(2021)Riquelme, Puigcerver, Mustafa, Neumann, Jenatton, Susano~Pinto, Keysers, and Houlsby]{riquelme2021scaling}
Carlos Riquelme, Joan Puigcerver, Basil Mustafa, Maxim Neumann, Rodolphe Jenatton, Andr{\'e} Susano~Pinto, Daniel Keysers, and Neil Houlsby.
\newblock Scaling vision with sparse mixture of experts.
\newblock \emph{Advances in Neural Information Processing Systems}, 34:\penalty0 8583--8595, 2021.

\bibitem[Roller et~al.(2021)Roller, Sukhbaatar, Weston, et~al.]{roller2021hash}
Stephen Roller, Sainbayar Sukhbaatar, Jason Weston, et~al.
\newblock Hash layers for large sparse models.
\newblock \emph{Advances in Neural Information Processing Systems}, 34:\penalty0 17555--17566, 2021.

\bibitem[Shao et~al.(2022)Shao, Zhang, Wang, Wei, and Xu]{shao2022spatial}
Zezhi Shao, Zhao Zhang, Fei Wang, Wei Wei, and Yongjun Xu.
\newblock Spatial-temporal identity: A simple yet effective baseline for multivariate time series forecasting.
\newblock In \emph{Proceedings of the 31st ACM International Conference on Information \& Knowledge Management}, pages 4454--4458, 2022.

\bibitem[Shazeer et~al.(2017)Shazeer, Mirhoseini, Maziarz, Davis, Le, Hinton, and Dean]{shazeer2017outrageously}
Noam Shazeer, Azalia Mirhoseini, Krzysztof Maziarz, Andy Davis, Quoc Le, Geoffrey Hinton, and Jeff Dean.
\newblock Outrageously large neural networks: The sparsely-gated mixture-of-experts layer.
\newblock \emph{arXiv preprint arXiv:1701.06538}, 2017.

\bibitem[Shen et~al.(2023)Shen, Hou, Zhou, Du, Longpre, Wei, Chung, Zoph, Fedus, Chen, et~al.]{shen2023flan}
Sheng Shen, Le~Hou, Yanqi Zhou, Nan Du, Shayne Longpre, Jason Wei, Hyung~Won Chung, Barret Zoph, William Fedus, Xinyun Chen, et~al.
\newblock Flan-moe: Scaling instruction-finetuned language models with sparse mixture of experts.
\newblock \emph{arXiv preprint arXiv:2305.14705}, 2023.

\bibitem[Wang et~al.(2022)Wang, Peng, Huang, Wang, Chen, and Xiao]{wang2022micn}
Huiqiang Wang, Jian Peng, Feihu Huang, Jince Wang, Junhui Chen, and Yifei Xiao.
\newblock Micn: Multi-scale local and global context modeling for long-term series forecasting.
\newblock In \emph{The Eleventh International Conference on Learning Representations}, 2022.

\bibitem[Wang et~al.(2023{\natexlab{a}})Wang, Liu, and Sun]{wang2023tlnets}
Wei Wang, Yang Liu, and Hao Sun.
\newblock Tlnets: Transformation learning networks for long-range time-series prediction.
\newblock \emph{arXiv preprint arXiv:2305.15770}, 2023{\natexlab{a}}.

\bibitem[Wang et~al.(2023{\natexlab{b}})Wang, Nie, Sun, Nguyen, Mulvey, and Poor]{wang2023st}
Zepu Wang, Yuqi Nie, Peng Sun, Nam~H Nguyen, John Mulvey, and H~Vincent Poor.
\newblock St-mlp: A cascaded spatio-temporal linear framework with channel-independence strategy for traffic forecasting.
\newblock \emph{arXiv preprint arXiv:2308.07496}, 2023{\natexlab{b}}.

\bibitem[Wen et~al.(2022)Wen, Zhou, Zhang, Chen, Ma, Yan, and Sun]{wen2022transformers}
Qingsong Wen, Tian Zhou, Chaoli Zhang, Weiqi Chen, Ziqing Ma, Junchi Yan, and Liang Sun.
\newblock Transformers in time series: A survey.
\newblock \emph{arXiv preprint arXiv:2202.07125}, 2022.

\bibitem[Wu et~al.(2021)Wu, Xu, Wang, and Long]{wu2021autoformer}
Haixu Wu, Jiehui Xu, Jianmin Wang, and Mingsheng Long.
\newblock Autoformer: Decomposition transformers with auto-correlation for long-term series forecasting.
\newblock \emph{Advances in Neural Information Processing Systems}, 34:\penalty0 22419--22430, 2021.

\bibitem[Wu et~al.(2023)Wu, Zhou, Long, and Wang]{wu_interpretable_2023}
Haixu Wu, Hang Zhou, Mingsheng Long, and Jianmin Wang.
\newblock Interpretable weather forecasting for worldwide stations with a unified deep model.
\newblock \emph{Nature Machine Intelligence}, 5\penalty0 (6):\penalty0 602--611, June 2023.
\newblock ISSN 2522-5839.
\newblock \doi{10.1038/s42256-023-00667-9}.
\newblock URL \url{https://doi.org/10.1038/s42256-023-00667-9}.

\bibitem[Xu et~al.(2023)Xu, Zeng, and Xu]{xu2023fits}
Zhijian Xu, Ailing Zeng, and Qiang Xu.
\newblock Fits: Modeling time series with $10 k $ parameters.
\newblock \emph{arXiv preprint arXiv:2307.03756}, 2023.

\bibitem[Yi et~al.(2023)Yi, Guo, Wei, Zhou, Wang, and Xu]{yi2023edgemoe}
Rongjie Yi, Liwei Guo, Shiyun Wei, Ao~Zhou, Shangguang Wang, and Mengwei Xu.
\newblock Edgemoe: Fast on-device inference of moe-based large language models.
\newblock \emph{arXiv preprint arXiv:2308.14352}, 2023.

\bibitem[Yuksel et~al.(2012)Yuksel, Wilson, and Gader]{6215056}
Seniha~Esen Yuksel, Joseph~N. Wilson, and Paul~D. Gader.
\newblock Twenty years of mixture of experts.
\newblock \emph{IEEE Transactions on Neural Networks and Learning Systems}, 23\penalty0 (8):\penalty0 1177--1193, 2012.
\newblock \doi{10.1109/TNNLS.2012.2200299}.

\bibitem[Zeng et~al.(2023)Zeng, Chen, Zhang, and Xu]{zeng2023transformers}
Ailing Zeng, Muxi Chen, Lei Zhang, and Qiang Xu.
\newblock Are transformers effective for time series forecasting?
\newblock In \emph{Proceedings of the AAAI conference on artificial intelligence}, volume~37, pages 11121--11128, 2023.

\bibitem[Zhang et~al.(2022)Zhang, Zhang, Cao, Bian, Yi, Zheng, and Li]{zhang2022less}
Tianping Zhang, Yizhuo Zhang, Wei Cao, Jiang Bian, Xiaohan Yi, Shun Zheng, and Jian Li.
\newblock Less is more: Fast multivariate time series forecasting with light sampling-oriented mlp structures.
\newblock \emph{arXiv preprint arXiv:2207.01186}, 2022.

\bibitem[Zhang and Yan(2022)]{zhang2022crossformer}
Yunhao Zhang and Junchi Yan.
\newblock Crossformer: Transformer utilizing cross-dimension dependency for multivariate time series forecasting.
\newblock In \emph{The Eleventh International Conference on Learning Representations}, 2022.

\bibitem[Zhou et~al.(2021)Zhou, Zhang, Peng, Zhang, Li, Xiong, and Zhang]{zhou2021informer}
Haoyi Zhou, Shanghang Zhang, Jieqi Peng, Shuai Zhang, Jianxin Li, Hui Xiong, and Wancai Zhang.
\newblock Informer: Beyond efficient transformer for long sequence time-series forecasting.
\newblock In \emph{Proceedings of the AAAI conference on artificial intelligence}, volume~35, pages 11106--11115, 2021.

\bibitem[Zhou et~al.(2022{\natexlab{a}})Zhou, Ma, Wen, Wang, Sun, and Jin]{zhou2022fedformer}
Tian Zhou, Ziqing Ma, Qingsong Wen, Xue Wang, Liang Sun, and Rong Jin.
\newblock Fedformer: Frequency enhanced decomposed transformer for long-term series forecasting.
\newblock In \emph{International Conference on Machine Learning}, pages 27268--27286. PMLR, 2022{\natexlab{a}}.

\bibitem[Zhou et~al.(2022{\natexlab{b}})Zhou, Zhu, Wang, Ma, Wen, Sun, and Jin]{zhou2022treedrnet}
Tian Zhou, Jianqing Zhu, Xue Wang, Ziqing Ma, Qingsong Wen, Liang Sun, and Rong Jin.
\newblock Treedrnet: a robust deep model for long term time series forecasting.
\newblock \emph{arXiv preprint arXiv:2206.12106}, 2022{\natexlab{b}}.

\bibitem[Zhu et~al.(2023)Zhu, Xiong, Wu, Nie, Zhang, and Yang]{zhu2023weather2k}
Xun Zhu, Yutong Xiong, Ming Wu, Gaozhen Nie, Bin Zhang, and Ziheng Yang.
\newblock Weather2k: A multivariate spatio-temporal benchmark dataset for meteorological forecasting based on real-time observation data from ground weather stations.
\newblock \emph{arXiv preprint arXiv:2302.10493}, 2023.

\end{thebibliography}

%%%%%%%%%%%%%%%%%%%%%%%%%%%%%%%%%%%%%%%%%%%%%%%%%%%%%%%%%%%%
\section*{Checklist}

% %%% BEGIN INSTRUCTIONS %%%
% The checklist follows the references. For each question, choose your answer from the three possible options: Yes, No, Not Applicable.  You are encouraged to include a justification to your answer, either by referencing the appropriate section of your paper or providing a brief inline description (1-2 sentences). 
% Please do not modify the questions.  Note that the Checklist section does not count towards the page limit. Not including the checklist in the first submission won't result in desk rejection, although in such case we will ask you to upload it during the author response period and include it in camera ready (if accepted).

% \textbf{In your paper, please delete this instructions block and only keep the Checklist section heading above along with the questions/answers below.}
% %%% END INSTRUCTIONS %%%

 \begin{enumerate}

 \item For all models and algorithms presented, check if you include:
 \begin{enumerate}
   \item A clear description of the mathematical setting, assumptions, algorithm, and/or model. \textbf{Yes}
   \item An analysis of the properties and complexity (time, space, sample size) of any algorithm. \textbf{Yes}
   \item (Optional) Anonymized source code, with specification of all dependencies, including external libraries. \textbf{Yes}
 \end{enumerate}

 \item For any theoretical claim, check if you include:
 \begin{enumerate}
   \item Statements of the full set of assumptions of all theoretical results. \textbf{Not Applicable}
   \item Complete proofs of all theoretical results. \textbf{Not Applicable}
   \item Clear explanations of any assumptions. \textbf{Not Applicable}   
 \end{enumerate}

 \item For all figures and tables that present empirical results, check if you include:
 \begin{enumerate}
   \item The code, data, and instructions needed to reproduce the main experimental results (either in the supplemental material or as a URL). \textbf{Yes}
   \item All the training details (e.g., data splits, hyperparameters, how they were chosen). \textbf{Yes}
         \item A clear definition of the specific measure or statistics and error bars (e.g., with respect to the random seed after running experiments multiple times). \textbf{Yes}
         \item A description of the computing infrastructure used. (e.g., type of GPUs, internal cluster, or cloud provider). \textbf{Yes}
 \end{enumerate}

 \item If you are using existing assets (e.g., code, data, models) or curating/releasing new assets, check if you include:
 \begin{enumerate}
   \item Citations of the creator If your work uses existing assets. \textbf{Yes}
   \item The license information of the assets, if applicable. \textbf{Not Applicable}
   \item New assets either in the supplemental material or as a URL, if applicable. \textbf{Not Applicable}
   \item Information about consent from data providers/curators. \textbf{Not Applicable}
   \item Discussion of sensible content if applicable, e.g., personally identifiable information or offensive content. \textbf{Not Applicable}
 \end{enumerate}

 \item If you used crowdsourcing or conducted research with human subjects, check if you include:
 \begin{enumerate}
   \item The full text of instructions given to participants and screenshots. \textbf{Not Applicable}
   \item Descriptions of potential participant risks, with links to Institutional Review Board (IRB) approvals if applicable. \textbf{Not Applicable}
   \item The estimated hourly wage paid to participants and the total amount spent on participant compensation. \textbf{Not Applicable}
 \end{enumerate}

 \end{enumerate}

\appendix
\thispagestyle{empty}
\onecolumn
\aistatstitle
{Mixture-of-Linear-Experts for Long-term Time Series Forecasting: \\
Supplementary Materials}

    \section*{Supplementary Materials Contents}  % Title for the appendix ToC
    \setcounter{tocdepth}{1}  % Show sections in the appendix ToC
    \startcontents[appendices]  % Starts appendix ToC
    \printcontents[appendices]{l}{1}{\setcounter{tocdepth}{2}}

\section{Time Stamp Embedding}\label{app:embedding_time_stamps}

In the preprocessing of date-time information for our datasets, various time-based features are extracted from the timestamp data. Each feature corresponds to a specific component of the timestamp, including the hour of the day, day of the week, day of the month, and day of the year. These features are each encoded into a range of $-0.5$ to $0.5$, with the transformation depending on the possible range of each feature. 

For instance, in the case of the day of the week, Monday is indexed as $0$ and Sunday as $6$. Each day of the week is then encoded to a value between $-0.5$ and $0.5$, with the transformation given by the formula
\[
\left(\frac{\text{index}}{6}\right) - 0.5
\]
Thus, Monday would be encoded as $-0.5$, Tuesday as approximately $-0.33$, Wednesday as approximately $-0.17$, and so on, with Sunday encoded as $0.5$.

\section{Toy Dataset Experiments}

\subsection{Dataset Description}\label{toy_dataset_description}

We construct a toy dataset that models periodic behaviors across different days of the week. Precisely, the dataset can be written as:
\[
y(t) = 
\begin{cases} 
y_{M-T}(t) + \text{noise}(t) &\text{Monday to Thursday} \\
y_{F-S}(t) + \text{noise}(t) &\text{Friday to Sunday}
\end{cases}
\]
where 
\[
    y_{M-T}(t) = 6 \sin\left(2\pi \frac{1}{24} t\right) + 20
\]
\[
y_{F-S}(t) = 6 \sin\left(2\pi \frac{1}{12} t\right) + 20
\]
\[
\text{noise}(t) \sim \mathcal{N}(0, 0.1^2)
\]

\subsection{More detailed results from the toy dataset experiments}

\begin{figure}[H]
\vspace{.3in}
    \centering
    \begin{subfigure}[t]{0.235\textwidth} %
        \centering
        \includeinkscape[width=\linewidth]{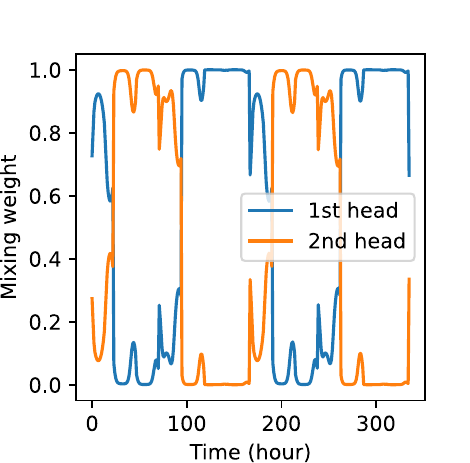}
        \caption{How mixing weights change with time}
        \label{fig:toy_dataset_details_1}
    \end{subfigure}
    \hfill
    \begin{subfigure}[t]{0.235\textwidth} %
        \centering
        \includeinkscape[width=\linewidth]{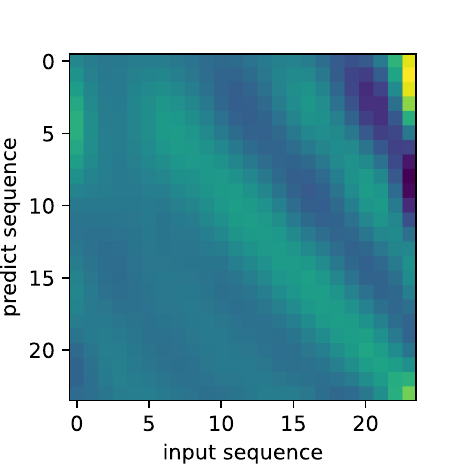}
        \caption{Weights of the linear layer in single-head model}
        \label{fig:toy_dataset_details_2}
    \end{subfigure}
    \vspace{1em} %
    \begin{subfigure}[t]{0.235\textwidth}
        \centering
        \includeinkscape[width=\linewidth]{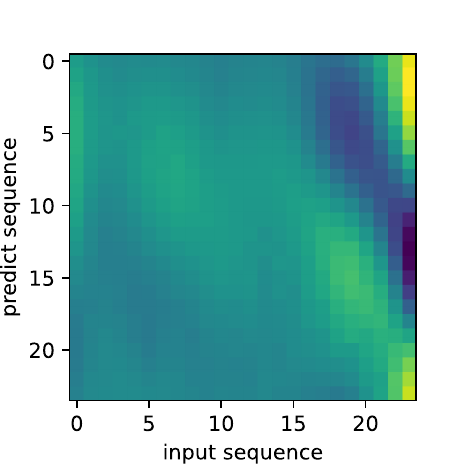}
        \caption{Weights of the linear layer in the 1st head of multi-head model}
        \label{fig:toy_dataset_details_3}
    \end{subfigure}
    \hfill
    \begin{subfigure}[t]{0.235\textwidth}
        \centering
        \includeinkscape[width=\linewidth]{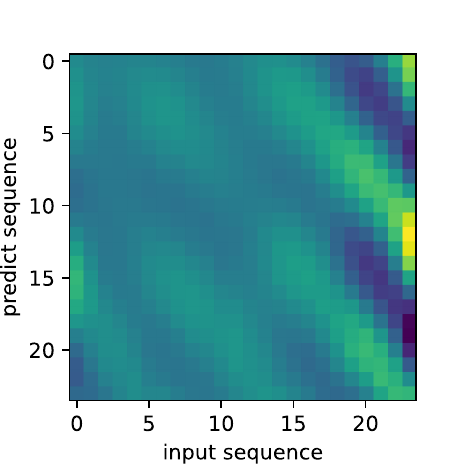}
        \caption{Weights of the linear layer in the 2nd head of multi-head model}
        \label{fig:toy_dataset_details_4}
    \end{subfigure}
    \vspace{.3in}
    \caption{More detailed look into the experiment of the toy examples. In heatmaps, a bright color indicates a positive value while a dark color indicates a negative value.}
    \label{fig:toy_dataset_details}
\end{figure}

Figure \ref{fig:toy_dataset_details_1} shows how the mixing weights change over a continuous 2-week period. The mixing layer captures the temporal patterns in the data, applying different heads for different frequencies. Figure \ref{fig:toy_dataset_details_2} is a heatmap representing the weight distribution of the linear layer in the single-head model. Using one head to predict the time series, the model derives a compromised weight set that underperforms, as shown in Figure \ref{fig:toy_dataset_output}. However, when employing two heads, the resulting models—whose linear layer weights are depicted in Figures \ref{fig:toy_dataset_details_3} and \ref{fig:toy_dataset_details_4}—efficiently learn the two frequency patterns and can yield more accurate predictions.

\section{Overview of Real-World Datasets}

Table \ref{tab:datasets} provides a brief overview of the real-world datasets evaluated. 

\begin{table}[H]
\caption{Overview of Datasets} \label{tab:datasets}
\begin{center}
\begin{tabular}{lll}
&\textbf{Size} &\textbf{Granu-}\\
\textbf{Dataset} &\textbf{(Timestamps}&\textbf{larity}\\
&\textbf{$\times$Channels)}&\\
\hline \\
ETTh1,h2 & 17420 (725.83 days) $\times$ 7 & 1 hour\\
ETTm1,m2 & 69680 (725.83 days) $\times$ 7 & 15 minutes\\
Weather & 52696 (365.86 days) $\times$ 21 & 10 minutes\\
Electricity & 26304 (1096 days) $\times$ 321 & 1 hour\\
Traffic & 17544 (731 days) $\times$ 336& 1 hour\\
Weather2K & 40896 (1704 days) $\times$ 20 & 1 hour
\end{tabular}
\end{center}
\end{table}

\section{Additional Main Results}
\label{app:main_results}

\subsection{Comparing Original and Our Replicated Experiments.}

In this section, we compare previously reported results with our reproduction of prior linear-centric results (using their code).

Table \ref{tab:reproduce} illustrates the differences between the original findings (Auth.) and our replicated experiments (Reprod.). On average, the DLinear model's outcomes vary by roughly 1.04\% from the original. Meanwhile, the RMLP and RLinear models show variations of around 2.10\% and 1.99\% respectively, indicating a generally consistent replication.

\begin{table*}[h]
\caption{Comparison of author's original results (Auth.) with reproduced results (Reprod.) Prediction length $\in \{96, 192, 336, 720\}$. The values reported are MSE loss. A lower value indicates a better prediction.  The Difference (Diff.) represents the relative differences from the original results, computed as $\text{Diff.} = (\text{Reprod.} - \text{Auth.})/{\text{Auth.}} \times 100\%$. 
% Blank entries indicate datasets and prediction lengths that were not reported by the authors.  
Entries marked with `--' indicate datasets and prediction lengths not reported in previous work.
} \label{tab:reproduce}
\begin{center}
\begin{tabular}{p{0.2cm}p{0.2cm}|}
\toprule
\multicolumn{2}{c}{\textbf{Model}}\\\cmidrule{1-2}
\\\midrule
\multirow{4}{*}{\rotatebox[origin=c]{90}{ETTh1}} &96 \\ &192\\ &336\\ &720\\\midrule
\multirow{4}{*}{\rotatebox[origin=c]{90}{ETTh2}} &96 \\ &192\\ &336\\ &720\\\midrule
\multirow{4}{*}{\rotatebox[origin=c]{90}{ETTm1}} &96 \\ &192\\ &336\\ &720\\\midrule
\multirow{4}{*}{\rotatebox[origin=c]{90}{ETTm2}} &96 \\ &192\\ &336\\ &720\\\midrule
\multirow{4}{*}{\rotatebox[origin=c]{90}{weather}} &96 \\ &192\\ &336\\ &720\\\midrule
\multirow{4}{*}{\rotatebox[origin=c]{90}{electricity}} &96 \\ &192\\ &336\\ &720\\\midrule
\multirow{4}{*}{\rotatebox[origin=c]{90}{traffic}} &96 \\ &192\\ &336\\ &720\\\midrule
\multicolumn{2}{c}{Avg. Diff.}\\
\bottomrule
\end{tabular}%
\begin{tabular}{ccc|ccc|ccc}\toprule
\multicolumn{3}{c}{DLinear} &\multicolumn{3}{c}{RLinear} &\multicolumn{3}{c}{RMLP} \\\cmidrule{1-9}
Auth. &Reprod. &Diff. &Auth. &Reprod. &Diff. &Auth. &Reprod. &Diff. \\\midrule
0.375 &0.372 &-0.80\% &0.366 &0.371 &1.37\% &0.390 &0.381 &-2.31\% \\
0.405 &0.413 &1.98\% &0.404 &0.404 &0.00\% &0.430 &0.541 &25.81\% \\
0.439 &0.442 &0.68\% &0.420 &0.428 &1.90\% &0.441 &0.453 &2.72\% \\
0.472 &0.501 &6.14\% &0.442 &0.45 &1.81\% &0.506 &0.502 &-0.79\% \\\midrule
0.289 &0.287 &-0.69\% &0.262 &0.272 &3.82\% &0.288 &0.294 &2.08\% \\
0.383 &0.349 &-8.88\% &0.319 &0.341 &6.90\% &0.343 &0.362 &5.54\% \\
0.448 &0.430 &-4.02\% &0.325 &0.372 &14.46\% &0.353 &0.389 &10.20\% \\
0.605 &0.710 &17.36\% &0.372 &0.418 &12.37\% &0.410 &0.440 &7.32\% \\\midrule
0.299 &0.300 &0.33\% &0.301 &0.301 &0.00\% &0.298 &0.300 &0.67\% \\
0.335 &0.336 &0.30\% &0.335 &0.335 &0.00\% &0.344 &0.339 &-1.45\% \\
0.369 &0.374 &1.36\% &0.370 &0.371 &0.27\% &0.390 &0.365 &-6.41\% \\
0.425 &0.461 &8.47\% &0.425 &0.429 &0.94\% &0.445 &0.439 &-1.35\% \\\midrule
0.167 &0.168 &0.60\% &0.164 &0.164 &0.00\% &0.174 &0.165 &-5.17\% \\
0.224 &0.228 &1.79\% &0.219 &0.219 &0.00\% &0.236 &0.223 &-5.51\% \\
0.281 &0.295 &4.98\% &0.273 &0.272 &-0.37\% &0.291 &0.282 &-3.09\% \\
0.397 &0.382 &-3.78\% &0.366 &0.368 &0.55\% &0.371 &0.362 &-2.43\% \\\midrule
0.176 &0.175 &-0.57\% &0.175 &0.174 &-0.57\% &0.149 &0.156 &4.70\% \\
0.220 &0.224 &1.82\% &0.218 &0.217 &-0.46\% &0.194 &0.203 &4.64\% \\
0.265 &0.263 &-0.75\% &0.265 &0.264 &-0.38\% &0.243 &0.254 &4.53\% \\
0.323 &0.332 &2.79\% &0.329 &0.331 &0.61\% &0.316 &0.331 &4.75\% \\\midrule
0.140 &0.140 &0.00\% &0.140 &0.143 &2.14\% &0.129 &0.131 &1.55\% \\
0.153 &0.153 &0.00\% &0.154 &0.157 &1.95\% &0.147 &0.149 &1.36\% \\
0.169 &0.169 &0.00\% &0.171 &0.174 &1.75\% &0.164 &0.167 &1.83\% \\
0.203 &0.203 &0.00\% &0.209 &0.212 &1.44\% &0.203 &0.200 &-1.48\% \\\midrule
0.410 &0.410 &0.00\% & -- &0.412 & -- & -- &0.380 & -- \\
0.423 &0.423 &0.00\% & -- &0.424 & -- & -- &0.396 & -- \\
0.436 &0.436 &0.00\% & -- &0.437 & -- & -- &0.409 & -- \\
0.466 &0.466 &0.00\% & -- &0.446 & -- & -- &0.441 & -- \\\midrule
& &1.04\% & & &2.10\% & & &1.99\% \\
\bottomrule
\end{tabular}
\end{center}
\end{table*}

\subsection{Results with Different Random Seeds}
Additionally, we conducted robustness experiments on all datasets except for Weather2K. Instead of utilizing a single seed (2021), we employed three different seeds (2021, 2022, 2023) and calculated the average losses from these runs.
Within each run, the reported losses still adhere to our primary experimental method where we chose the test losses of hyperparameters that resulted in the lowest validation losses.

Table \ref{tab:robustness} presents the results of our robustness tests. Here, MoLE exhibited improvements over the DLinear model in 16 settings, and over RLinear and RMLP in 21 and 19 settings, respectively. The average enhancements exceed 67.7\%. This demonstrates that our methodology remains consistent across varied random seeds and consistently enhances the performance of the linear-centric models we tested.

\begin{table*}[h]
\caption{Comparison between original (single-head) and enhanced (multi-head \nameshort{}) DLinear, RLinear, and RMLP models. Each experiment is conducted using three different random seeds: 2021, 2022, 2023. The average losses (avg) and standard deviations (stdev) are reported. Prediction length $\in \{96, 192, 336, 720\}$. The values reported are MSE loss. A lower value indicates a better prediction. The cells shown in \textcolor{blue}{blue} indicate that \nameshort{} improves the performance of the original (single-head) models. 
} \label{tab:robustness}
\begin{center}
\resizebox{\textwidth}{!}{
\begin{tabular}{p{0.2cm}p{0.2cm}|}
\toprule
\multicolumn{2}{c}{\textbf{Model}}\\\cmidrule{1-2}
\\\cmidrule{1-2}
\\\midrule
\multirow{4}{*}{\rotatebox[origin=c]{90}{ETTh1}} &96 \\ &192\\ &336\\ &720\\\midrule
\multirow{4}{*}{\rotatebox[origin=c]{90}{ETTh2}} &96 \\ &192\\ &336\\ &720\\\midrule
\multirow{4}{*}{\rotatebox[origin=c]{90}{ETTm1}} &96 \\ &192\\ &336\\ &720\\\midrule
\multirow{4}{*}{\rotatebox[origin=c]{90}{ETTm2}} &96 \\ &192\\ &336\\ &720\\\midrule
\multirow{4}{*}{\rotatebox[origin=c]{90}{weather}} &96 \\ &192\\ &336\\ &720\\\midrule
\multirow{4}{*}{\rotatebox[origin=c]{90}{electricity}} &96 \\ &192\\ &336\\ &720\\\midrule
\multirow{4}{*}{\rotatebox[origin=c]{90}{traffic}} &96 \\ &192\\ &336\\ &720\\\midrule
\multicolumn{2}{c}{IMP}\\
\bottomrule
\end{tabular}%
\begin{tabular}{cccc|cccc|cccc}\toprule
    \multicolumn{4}{c}{DLinear} &\multicolumn{4}{c}{RLinear} &\multicolumn{4}{c}{RMLP} \\\cmidrule{1-12}
    \multicolumn{2}{c}{original} &\multicolumn{2}{c|}{\nameshort{}} &\multicolumn{2}{c}{original} &\multicolumn{2}{c|}{\nameshort{}} &\multicolumn{2}{c}{original} &\multicolumn{2}{c}{\nameshort{}} \\\cmidrule{1-12}
    avg &stdev &avg &stdev &avg &stdev &avg &stdev &avg &stdev &avg &stdev \\\midrule
    0.3741 &2.29e-3 &0.3824 &1.58e-2 &0.3711 &1.67e-3 &0.3742 &2.12e-3 &0.3833 &2.20e-3 &0.4109 &7.29e-3 \\
    0.4134 &7.83e-3 &0.4291 &2.53e-2 &0.4052 &1.60e-3 &0.4105 &6.61e-3 &0.4914 &4.75e-2 &\color{blue}0.4403 &1.06e-2 \\
    0.4499 &1.14e-2 &0.4557 &2.16e-3 &0.4291 &1.52e-3 &0.5252 &1.67e-1 &0.5028 &2.86e-2 &\color{blue}0.4433 &8.53e-3 \\
    0.5072 &1.76e-2 &0.5136 &3.02e-2 &0.4483 &4.01e-4 &0.4493 &2.18e-4 &0.5072 &1.54e-2 &\color{blue}0.4830 &1.19e-2 \\\midrule
    0.2824 &4.96e-3 &0.2842 &3.68e-3 &0.2730 &2.02e-3 &0.2737 &6.71e-4 &0.2943 &4.31e-4 &0.2979 &7.74e-3 \\
    0.3566 &6.84e-3 &0.3652 &2.32e-2 &0.3471 &5.95e-3 &\color{blue}0.3401 &2.40e-3 &0.3683 &9.96e-3 &0.4028 &7.25e-2 \\
    0.4162 &2.48e-2 &0.4191 &2.62e-3 &0.3754 &3.47e-3 &0.3770 &7.80e-3 &0.3926 &1.24e-2 &0.4044 &1.02e-2 \\
    0.6648 &5.46e-2 &\color{blue}0.6238 &8.74e-2 &0.4233 &3.55e-3 &\color{blue}0.3918 &6.08e-3 &0.4444 &1.61e-2 &\color{blue}0.4206 &9.33e-3 \\\midrule
    0.2998 &3.31e-4 &\color{blue}0.2879 &1.14e-3 &0.3024 &1.73e-3 &\color{blue}0.2925 &4.47e-3 &0.2968 &3.58e-3 &0.2983 &1.86e-3 \\
    0.3358 &2.84e-4 &\color{blue}0.3349 &7.72e-3 &0.3402 &3.79e-3 &\color{blue}0.3313 &2.12e-3 &0.3365 &3.25e-3 &\color{blue}0.3348 &3.32e-3 \\
    0.3794 &5.20e-3 &\color{blue}0.3731 &4.19e-3 &0.3714 &8.36e-4 &\color{blue}0.3692 &1.82e-3 &0.3729 &3.30e-3 &\color{blue}0.3697 &5.22e-3 \\
    0.4400 &1.13e-3 &0.4486 &1.82e-2 &0.4274 &2.05e-3 &\color{blue}0.4271 &9.96e-4 &0.4306 &2.64e-3 &\color{blue}0.4284 &3.20e-3 \\\midrule
    0.1680 &9.57e-4 &0.1684 &2.80e-3 &0.1636 &9.65e-4 &\color{blue}0.1622 &6.40e-4 &0.1672 &3.59e-3 &0.1672 &4.86e-3 \\
    0.2291 &7.67e-4 &\color{blue}0.2286 &3.18e-3 &0.2195 &7.68e-4 &\color{blue}0.2170 &3.70e-4 &0.2216 &2.35e-3 &\color{blue}0.2190 &1.20e-3 \\
    0.2875 &9.00e-3 &0.2956 &5.78e-3 &0.2724 &3.48e-4 &\color{blue}0.2720 &7.32e-4 &0.2807 &5.57e-3 &\color{blue}0.2792 &4.38e-3 \\
    0.3990 &1.73e-2 &0.4082 &1.29e-3 &0.3678 &3.03e-4 &0.3688 &2.16e-3 &0.3680 &4.33e-3 &0.3729 &6.62e-3 \\\midrule
    0.1748 &2.05e-4 &\color{blue}0.1599 &2.33e-2 &0.1742 &6.24e-5 &\color{blue}0.1506 &4.39e-3 &0.1560 &1.99e-3 &\color{blue}0.1549 &9.27e-3 \\
    0.2182 &4.63e-3 &\color{blue}0.2014 &9.50e-3 &0.2166 &6.24e-4 &\color{blue}0.1890 &7.38e-4 &0.2037 &1.67e-3 &\color{blue}0.1952 &5.40e-3 \\
    0.2656 &5.52e-3 &0.3201 &1.33e-1 &0.2640 &2.34e-4 &\color{blue}0.2449 &5.00e-3 &0.2564 &2.49e-3 &\color{blue}0.2464 &1.79e-3 \\
    0.3295 &3.03e-3 &\color{blue}0.3125 &1.93e-3 &0.3311 &6.52e-5 &\color{blue}0.3204 &5.97e-3 &0.3269 &2.52e-3 &0.3931 &1.06e-1 \\\midrule
    0.1399 &4.70e-6 &\color{blue}0.1289 &3.64e-4 &0.1432 &2.74e-5 &\color{blue}0.1319 &1.72e-4 &0.1307 &6.23e-4 &\color{blue}0.1290 &6.40e-4 \\
    0.1533 &1.75e-6 &\color{blue}0.1475 &2.90e-4 &0.1568 &1.79e-6 &\color{blue}0.1513 &4.82e-4 &0.1497 &2.62e-4 &0.1511 &1.73e-3 \\
    0.1689 &6.38e-6 &\color{blue}0.1611 &1.26e-3 &0.1735 &5.27e-6 &\color{blue}0.1647 &1.04e-3 &0.1660 &6.25e-4 &\color{blue}0.1657 &2.61e-3 \\
    0.2032 &6.36e-5 &\color{blue}0.1787 &1.06e-3 &0.2122 &6.06e-5 &\color{blue}0.1842 &4.38e-3 &0.2022 &1.11e-3 &\color{blue}0.1795 &9.70e-4 \\\midrule
    0.4102 &2.60e-5 &\color{blue}0.3843 &1.25e-3 &0.4119 &6.46e-6 &\color{blue}0.3783 &2.29e-3 &0.3786 &2.70e-3 &\color{blue}0.3695 &3.48e-3 \\
    0.4226 &3.88e-5 &\color{blue}0.3957 &4.27e-3 &0.4243 &2.42e-5 &\color{blue}0.3949 &3.21e-3 &0.3967 &5.31e-4 &\color{blue}0.3852 &2.46e-3 \\
    0.4355 &3.40e-5 &\color{blue}0.4239 &3.08e-3 &0.4367 &2.14e-5 &\color{blue}0.4158 &2.31e-3 &0.4092 &1.13e-4 &\color{blue}0.4080 &5.02e-3 \\
    0.4657 &8.93e-5 &\color{blue}0.4533 &4.88e-3 &0.4655 &4.47e-6 &\color{blue}0.4538 &2.25e-3 &0.4436 &1.59e-3 &\color{blue}0.4408 &1.27e-3 \\\midrule
    \multicolumn{4}{c}{16 (57.1\%)} &\multicolumn{4}{c}{21 (75.0\%)} &\multicolumn{4}{c}{19 (67.9\%)}\\
    \bottomrule
    \end{tabular}
}
\end{center}
\end{table*}

\subsection{Results with Additional Hyperparameter Tuning}
\label{app:full_hp_tuning}

We are interested in understanding the comprehensive impact of batch sizes and head dropout rates on model performance. To explore this, we have added additional values of these hyperparameters into our hyperparameter tuning process. Due to the high costs of these experiments and their similar performance compared to the hyperparameter tuning shown in our main text, their results are presented only here in the Appendix. The grid search included the following sets of hyperparameters in Table \ref{tab:hyperparameters_appendix}.

Our methodology mirrors that of our primary experiments, where test losses correspond to hyperparameters yielding the lowest validation losses. Moreover, each experiment was conducted three times using distinct random seeds (2021, 2022, and 2023). The best results from each seed were averaged to obtain the final reported outcomes. The results in Table \ref{tab:dlinear_appendix} show that adding new hyperparameter values did not improve model performance. This suggests that the original set of hyperparameters was sufficient for representing each model's capabilities.

\begin{table}[H]
\caption{Hyperparameter search values. \textbf{Bold} denotes newly added hyperparameter values.} \label{tab:hyperparameters_appendix}
\begin{center}
\begin{tabular}{ll}
\textbf{Hyperparameter}  &\textbf{Values} \\
\hline \\
Batch size         &8, \textbf{16, 32, 64, 128} \\
Initial learning rate             &0.005, 0.01, 0.05 \\
Number of heads             &  2, 3, 4, 5, 6\\
Head dropout rate & 0, 0.2, \textbf{0.4, 0.6, 0.8} \\
Input sequence length & 336 \\
Prediction length &  \{96,192,336,720\}
\end{tabular}
\end{center}
\end{table}

\begin{table}[H]
\caption{Comparison between original (single-head) and enhanced (multi-head \nameshort{}) DLinear, RLinear, and RMLP models with added hyperparameter values. Prediction length $\in \{96, 192, 336, 720\}$. The values reported are MSE loss. A lower value indicates a better prediction. The better results among a linear-centric method and its \nameshort{} variant are highlighted in \textbf{bold.} 
} 
\label{tab:dlinear_appendix}
\begin{center}
\begin{tabular}{lc|cc|cc|cc}
\toprule
\multicolumn{2}{c}{\textbf{Model}} &\multicolumn{2}{c}{DLinear} &\multicolumn{2}{c}{RLinear} &\multicolumn{2}{c}{RMLP} \\\cmidrule{1-8}
\multicolumn{2}{c}{Dataset ~~ Prediction length} &original &\nameshort{} &original &\nameshort{} &original &\nameshort{} \\\midrule

\multirow{4}{*}{ETTh1} &96 &0.376 &\textbf{0.375} &\textbf{0.370} &0.374 &0.387 &\textbf{0.381} \\
&192 &\textbf{0.411} &0.436 &\textbf{0.405} &0.408 &0.434 &\textbf{0.425} \\
&336 &\textbf{0.456} &0.474 &\textbf{0.427} &0.433 &0.462 &\textbf{0.437} \\
&720 &0.498 &\textbf{0.497} &0.447 &\textbf{0.446} &\textbf{0.514} &0.524 \\\midrule
\multirow{4}{*}{ETTh2} &96 &\textbf{0.282} &\textbf{0.282} &\textbf{0.275} &\textbf{0.275} &0.304 &\textbf{0.285} \\
&192 &0.362 &\textbf{0.359} &\textbf{0.349} &\textbf{0.349} &\textbf{0.371} &0.372 \\
&336 &\textbf{0.403} &\textbf{0.403} &0.378 &\textbf{0.361} &0.399 &\textbf{0.398} \\
&720 &\textbf{0.610} &1.378 &0.423 &\textbf{0.389} &0.448 &\textbf{0.418} \\\midrule
\multirow{4}{*}{ETTm1} &96 &0.301 &\textbf{0.295} &0.304 &\textbf{0.290} &0.302 &\textbf{0.296} \\
&192 &\textbf{0.344} &0.347 &0.341 &\textbf{0.339} &0.341 &\textbf{0.334} \\
&336 &\textbf{0.382} &0.384 &0.374 &\textbf{0.372} &\textbf{0.373} &0.374 \\
&720 &\textbf{0.451} &0.466 &\textbf{0.428} &0.429 &0.432 &\textbf{0.428} \\\midrule
\multirow{4}{*}{ETTm2} &96 &0.166 &\textbf{0.165} &0.164 &\textbf{0.162} &0.170 &\textbf{0.169} \\
&192 &0.228 &\textbf{0.223} &\textbf{0.220} &\textbf{0.220} &\textbf{0.222} &0.228 \\
&336 &0.279 &\textbf{0.276} &0.273 &\textbf{0.272} &0.281 &\textbf{0.273} \\
&720 &0.392 &\textbf{0.386} &\textbf{0.367} &0.370 &\textbf{0.369} &0.371 \\\midrule
\multirow{4}{*}{weather} &96 &0.176 &\textbf{0.160} &0.174 &\textbf{0.149} &\textbf{0.152} &\textbf{0.152} \\
&192 &0.217 &\textbf{0.210} &0.217 &\textbf{0.195} &\textbf{0.194} &\textbf{0.194} \\
&336 &\textbf{0.266} &0.300 &0.265 &\textbf{0.243} &\textbf{0.244} &0.246 \\
&720 &0.334 &\textbf{0.332} &0.331 &\textbf{0.316} &0.325 &\textbf{0.322} \\\midrule
\multirow{4}{*}{electricity} &96 &0.140 &\textbf{0.130} &0.143 &\textbf{0.132} &0.130 &\textbf{0.128} \\
&192 &0.153 &\textbf{0.147} &0.154 &\textbf{0.149} &\textbf{0.148} &0.149 \\
&336 &0.169 &\textbf{0.160} &0.172 &\textbf{0.164} &0.164 &\textbf{0.163} \\
&720 &0.203 &\textbf{0.180} &0.211 &\textbf{0.189} &0.202 &\textbf{0.178} \\\midrule
\multirow{4}{*}{traffic} &96 &0.409 &\textbf{0.384} &0.412 &\textbf{0.383} &0.376 &\textbf{0.367} \\
&192 &0.420 &\textbf{0.400} &0.423 &\textbf{0.399} &0.396 &\textbf{0.385} \\
&336 &0.436 &\textbf{0.419} &0.437 &\textbf{0.413} &0.409 &\textbf{0.402} \\
&720 &0.466 &\textbf{0.448} &0.465 &\textbf{0.444} &0.443 &\textbf{0.435} \\\midrule

\multicolumn{2}{c}{No. improved (\%)} &\multicolumn{2}{c}{19 (67.9\%)} &\multicolumn{2}{c}{20 (71.4\%)} &\multicolumn{2}{c}{19 (67.9\%)}\\

\bottomrule
\end{tabular}
\end{center}
\end{table}

\section{Additional Ablations}

Based on the results of the full hyperparameter tuning shown in Appendix \ref{app:full_hp_tuning}, we conducted two additional ablations to further discuss how the choices of batch sizes and head dropout rates impact model performance. The plots that follow show curves which are the average test losses from three different seeds. These test losses are chosen from the hyperparameter configurations that yielded the lowest validation losses, while keeping the parameter on the x-axis constant. The shaded regions around each curve indicate the range of \(\pm\) one standard deviation.

\subsection{Impact of Batch Sizes on Model Performance}
\label{app:batch_size}

\begin{figure}[H]
    \centering
    \includeinkscape[width=\textwidth]{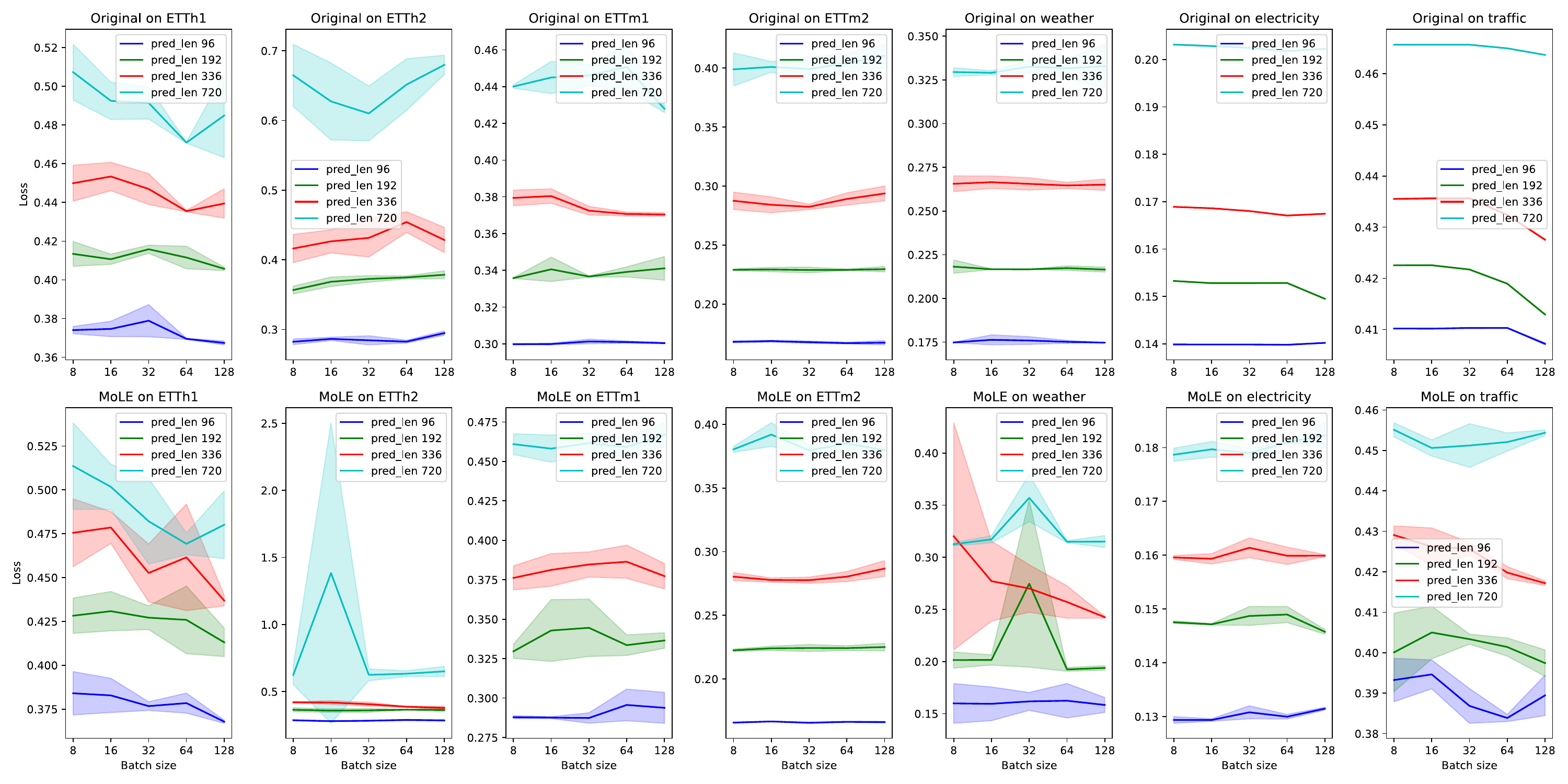}
    \caption{Comparison of test loss against batch sizes for DLinear/\nameshort{-DLinear} models across various prediction lengths and datasets. Top row: Single-head (Original) models. Bottom row: \nameshort{} models.}
    \label{fig:batch_size_appendix_dlinear}
\end{figure}

\begin{figure}[H]
    \centering
    \includeinkscape[width=\textwidth]{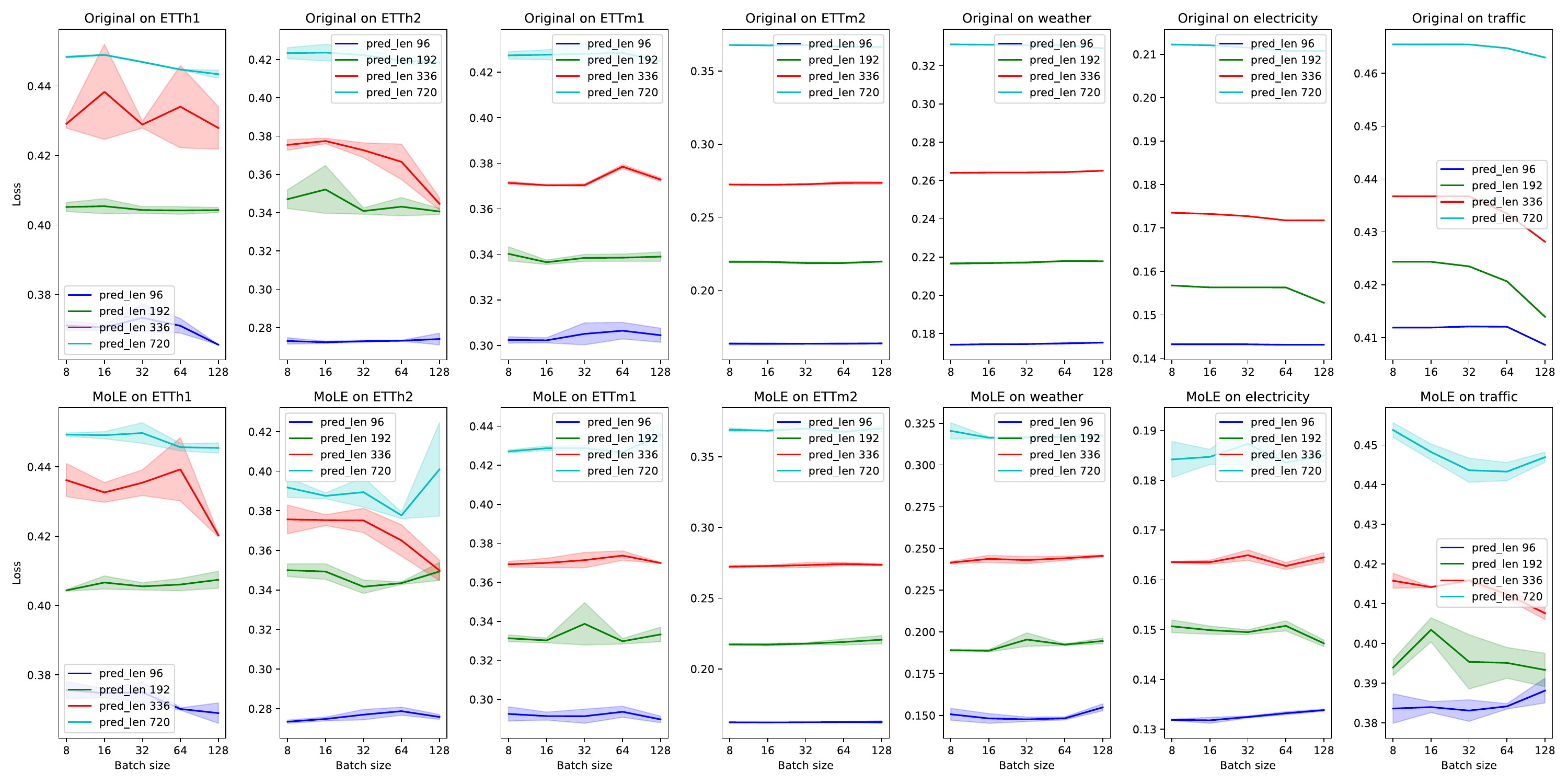}
    \caption{Comparison of test loss against batch sizes for RLinear/\nameshort{-RLinear} models across various prediction lengths and datasets. Top row: Single-head (Original) models. Bottom row: \nameshort{} models.}
    \label{fig:batch_size_appendix_rlinear}
\end{figure}

\begin{figure}[H]
    \centering
    \includeinkscape[width=\textwidth]{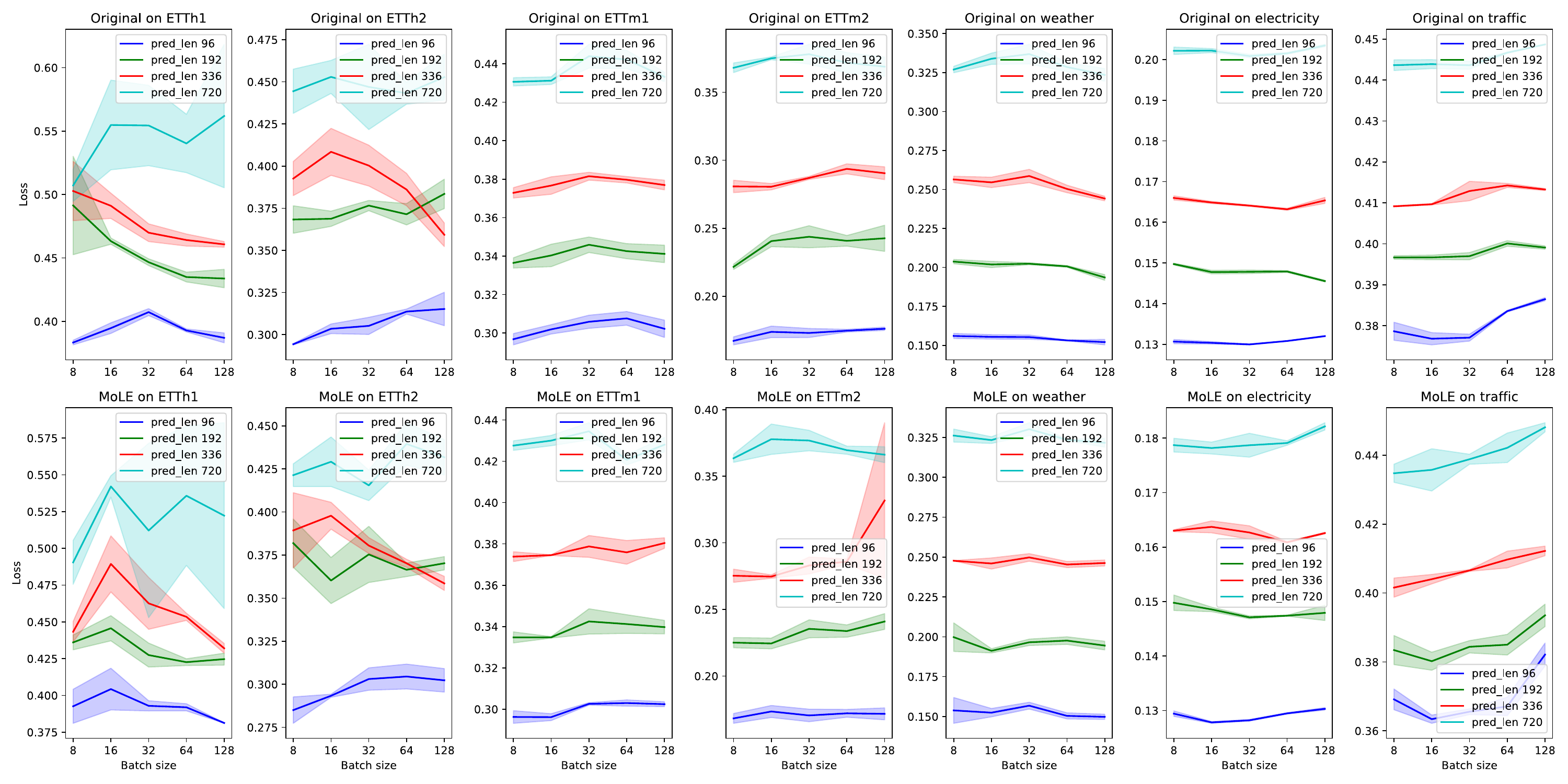}
    \caption{Comparison of test loss against batch sizes for RMLP/\nameshort{-RMLP} models across various prediction lengths and datasets. Top row: Single-head (Original) models. Bottom row: \nameshort{} models.}
    \label{fig:batch_size_appendix_rmlp}
\end{figure}

Figures \ref{fig:batch_size_appendix_dlinear}, \ref{fig:batch_size_appendix_rlinear} and \ref{fig:batch_size_appendix_rmlp} demonstrate the impact of varying batch sizes on the performance of both single-head and \nameshort{} DLinear, RLinear, and RMLP models. For smaller datasets, such as the ETT series, there is a trend similar to an inverse U-shape, suggesting that very small or very large batch sizes may enhance model performance. In contrast, for larger datasets, particularly electricity and traffic, larger batch sizes appear to benefit single-head models, while \nameshort{} models show optimal performance with mid-range to small batch sizes. Importantly, in large datasets, \nameshort{} models consistently outperform single-head models, even when the latter are given the advantage of larger batch sizes. This trend is also observable in Table \ref{tab:dlinear_appendix}, where comprehensive hyperparameter tuning does not enable single-head models to outperform \nameshort{} in large datasets. These results confirm that the batch size of 8, used in our main experiments, is sufficient to effectively represent each model's capabilities.

\subsection{Impact of Head Dropout Rates on \nameshort{} Performance}
\label{app:dropout_rate}

\begin{figure}
  \centering
  
  \begin{subfigure}[b]{\textwidth}
    \includeinkscape[width=\textwidth]{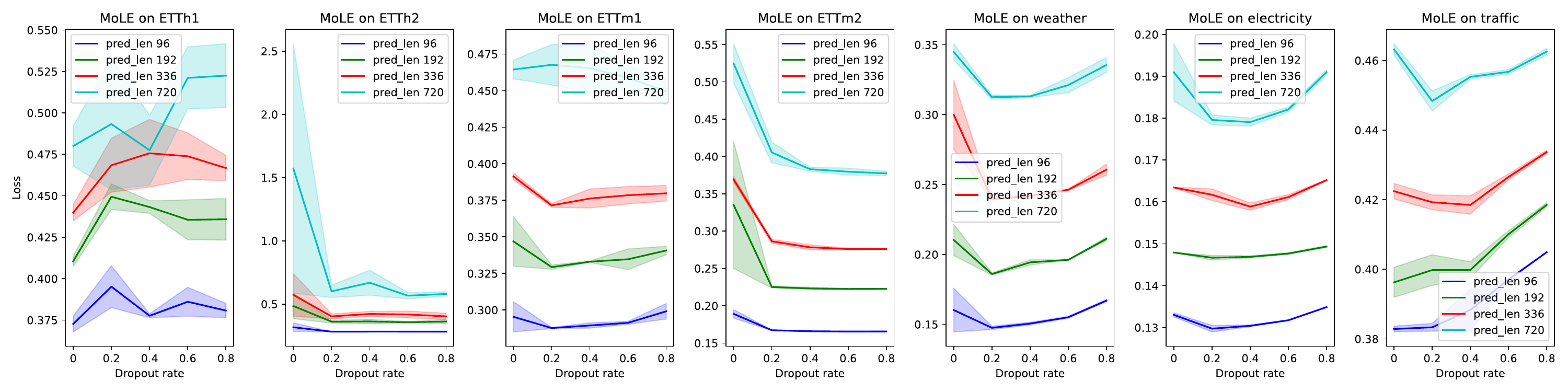}
    \caption{DLinear}
  \end{subfigure}
  \hfill %
  
  \begin{subfigure}[b]{\textwidth}
    \includeinkscape[width=\textwidth]{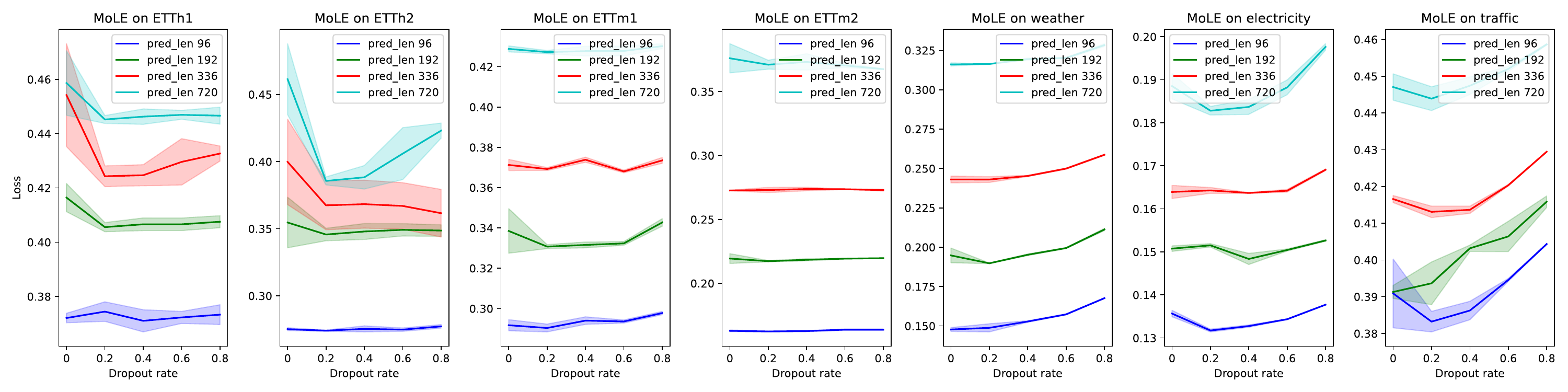}
    \caption{RLinear}
  \end{subfigure}
  \hfill
  
  \begin{subfigure}[b]{\textwidth}
    \includeinkscape[width=\textwidth]{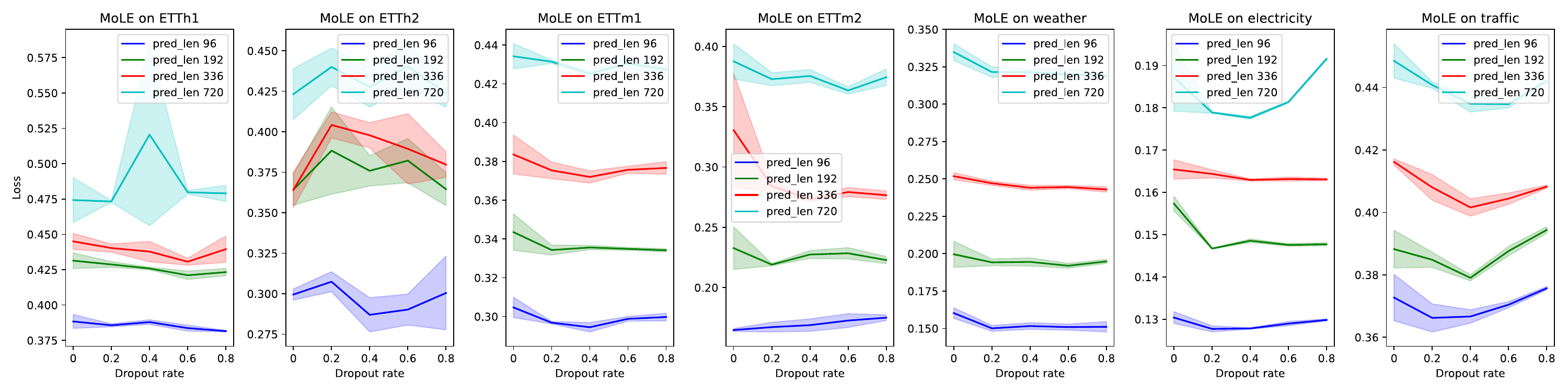}
    \caption{RMLP}
  \end{subfigure}
  
  \caption{Subplot comparison of test loss against head dropout rates for MoLE models across various prediction lengths and datasets.}
  \label{fig:dropout_appendix}
\end{figure}

Figure \ref{fig:dropout_appendix} illustrates the test loss as a function of dropout rates for three distinct models: DLinear, RLinear, and RMLP, across various datasets.

For the DLinear model, there is a consistent trend across datasets where certain mid-range dropout rates yield lower test losses, such as in the ETTm1, weather, electricity, and traffic datasets.
For RLinear and RMLP, similar U-shaped patterns can be observed, but only in larger datasets such as weather, electricity, and traffic.
For smaller datasets, like the ETT series, the link between dropout rate and test loss isn't clear. The plots show a lot of variation, indicating that the relationship is not strong.

% \begin{figure}[h]
%   \centering
  
%   \begin{subfigure}[b]{0.85\textwidth}
%     \includeinkscape[width=\textwidth]{svg-inkscape/batch_size_DLinear_svg-tex.pdf_tex}
%     \caption{DLinear}
%   \end{subfigure}
  
%   \begin{subfigure}[b]{0.85\textwidth}
%     \includeinkscape[width=\textwidth]{svg-inkscape/batch_size_RLinear_svg-tex.pdf_tex}
%     \caption{RLinear}
%   \end{subfigure}
  
%   \begin{subfigure}[b]{0.85\textwidth}
%     \includeinkscape[width=\textwidth]{svg-inkscape/batch_size_RMLP_svg-tex.pdf_tex}
%     \caption{RMLP}
%   \end{subfigure}
  
%   \caption{Subplot comparison of test loss versus batch sizes across different prediction lengths and datasets. First row of each subplot: single-head (Original) models. Second row: MoLE models.}
%   \label{fig:batch_size_appendix}
% \end{figure}

\filbreak
\section{Runtime Analysis}

Here, we examine how runtime performance is affected by employing multiple heads (\nameshort{}, our method) in contrast to a single-head configuration (original model). The experiments were conducted on a platform equipped with an NVIDIA Tesla V100-32GB SXM2 GPU and two shared Intel Xeon Gold 6248 CPUs. The ETTh1 dataset was used for this runtime analysis, with hyperparameters specified in Table \ref{tab:hyperparameter_runtime}.

\begin{wraptable}{r}{0.5\textwidth} % Aligns the table to the right; adjust the width as needed
\centering
\caption{Hyperparameter settings for runtime analysis}
\label{tab:hyperparameter_runtime}
\begin{tabular}{l l}
\hline
\textbf{Hyperparameter} & \textbf{Value} \\
\hline
Batch size & 8 \\
Initial learning rate & 0.005 \\
Head dropout rate & 0\\
Input sequence length & 336\\
Prediction length & 336\\
\hline
\end{tabular}
\end{wraptable}

\begin{figure}[H]
    \caption*{Training Time per iteration by Number of Heads}
    \centering
    % First row of subplots
    \begin{subfigure}{.33\linewidth}
        \centering
        \includegraphics[width=\linewidth]{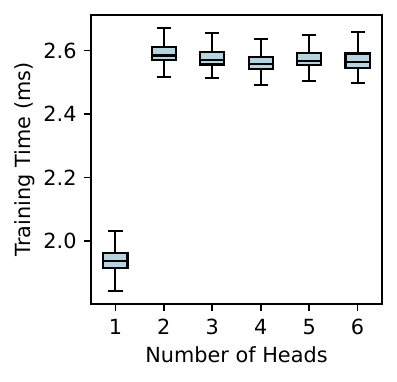}
        \caption{\nameshort{-DLinear} - Training}
    \end{subfigure}%
    \begin{subfigure}{.33\linewidth}
        \centering
        \includegraphics[width=\linewidth]{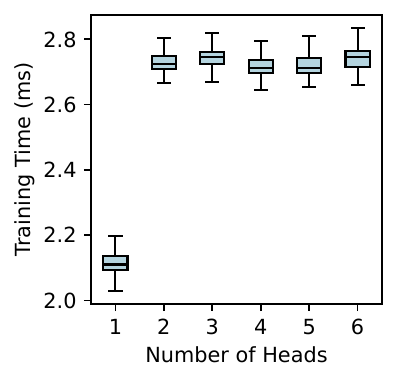}
        \caption{\nameshort{-RLinear} - Training}
    \end{subfigure}%
    \begin{subfigure}{.33\linewidth}
        \centering
        \includegraphics[width=\linewidth]{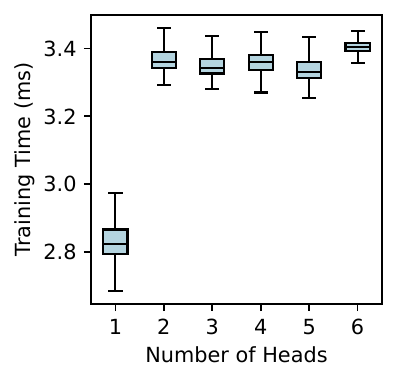}
        \caption{\nameshort{-RMLP} - Training}
    \end{subfigure}

    % Second row of subplots
    \caption*{Inference Time per iteration by Number of Heads}
    \begin{subfigure}{.33\linewidth}
        \centering
        \includegraphics[width=\linewidth]{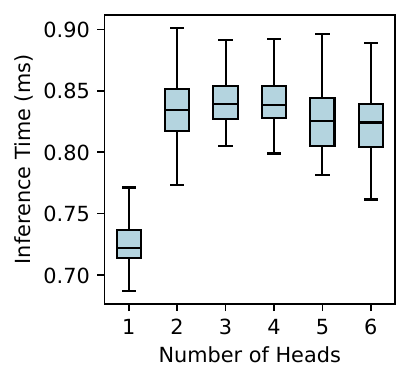}
        \caption{\nameshort{-DLinear} - Inference}
    \end{subfigure}%
    \begin{subfigure}{.33\linewidth}
        \centering
        \includegraphics[width=\linewidth]{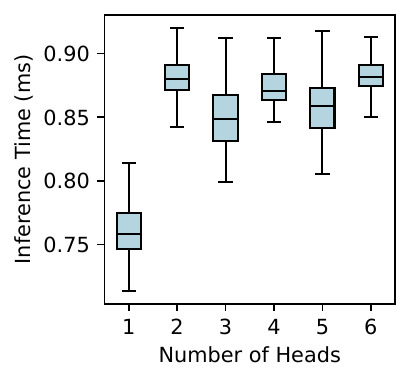}
        \caption{\nameshort{-RLinear} - Inference}
    \end{subfigure}%
    \begin{subfigure}{.33\linewidth}
        \centering
        \includegraphics[width=\linewidth]{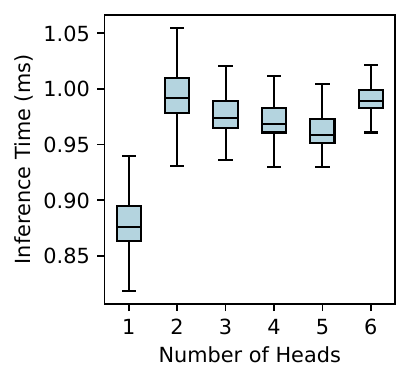}
        \caption{\nameshort{-RMLP} - Inference}
    \end{subfigure}

    \caption{Box plots illustrating the comparison of per-iteration training and inference times between original single-head and multi-head \nameshort{} models. Refer to footnote \ref{footnote:boxplot} for an explanation of the plot components.}
    \label{fig:runtime}
\end{figure}

Table \ref{tab:runtime} presents a comparison of training and inference times per iteration, as well as the number of parameters across different numbers of heads in each linear-centric model, including DLinear, RLinear, and RMLP. For single-head models, the original implementations are used, excluding the mixing layer present in \nameshort{}. 
Figure \ref{fig:runtime} presents box plots\footnote{\label{footnote:boxplot}
A box plot displays data distribution using five key metrics: the minimum, first quartile (Q1), median, third quartile (Q3), and the maximum. The box spans from Q1 to Q3, indicating the interquartile range (IQR), with a line inside marking the median. `Whiskers' extend from the box to the maximum and minimum values within 1.5*IQR from Q1 and Q3, and points outside this range are outliers and are not displayed.
} illustrating the medians and quartiles of training and inference times. The timing data was recorded at each iteration during the training and testing phases.

Table \ref{tab:runtime} and Figure \ref{fig:runtime} clearly demonstrate that the time overhead in \nameshort{} is attributable mostly to the additional mixing layer and the mixing mechanism, rather than the computation associated with extra heads. This efficiency is achieved through an implementation strategy that multiplies the output size of the main linear layer in the model by the number of heads, without adding extra layers. In comparison to the original single-head model, \nameshort{} introduces an approximate 26.45\% time overhead during training and 13.34\% during inference, irrespective of the number of heads utilized. However, it is important to note that, inherent to the design of Mixture of Experts (MoE) methods, the number of parameters in the model increases approximately proportionally with the number of heads.

\begin{table}[H]
\centering
\caption{Comparison of average training and inference times per iteration, along with the total number of parameters, for \nameshort{-DLinear}, \nameshort{-RLinear}, and \nameshort{-RMLP} models employing 1 to 6 heads. For single-head configurations, models are evaluated without the mixing layer and the mixing mechanism. ``Train. Time'' denotes average training time per iteration, ``Infer. Time'' represents average inference time per iteration, and ``Num. of Parameters'' indicates the total parameter count.}
\label{tab:runtime}
\begin{tabular}{ll|ccccccc}\toprule
\multirow{2}{*}{\textbf{Model}} &\multirow{2}{*}{\textbf{Metric}} &\multicolumn{6}{c}{\textbf{Num. of heads}} \\\cmidrule{3-8}
% & &Original &\multicolumn{5}{c}{MoLE} \\\cmidrule{3-8}
& &1 (original) &2 &3 &4 &5 &6 \\\midrule
\multirow{3}{*}{DLinear} & Train. Time (ms) & 1.950 & 2.604 & 2.594 & 2.571 & 2.583 & 2.584\\
&Infer. Time (ms) & 0.727 & 0.840 & 0.843 & 0.844 & 0.828 & 0.234\\
&Num. of Parameters & 226,464 & 453,208 & 679,959 & 906,808 & 1133,755 & 1360,800\\\midrule
\multirow{3}{*}{RLinear} &Train. Time (ms) & 2.129 & 2.737 & 2.751 & 2.727 & 2.730 & 2.752 \\
&Infer. Time (ms) & 0.763 & 0.882 & 0.851 & 0.879 & 0.858 & 0.885 \\
&Num. of Parameters & 113,246 & 226,758 & 340,277 & 453,894 & 567,609 & 681,422\\\midrule
\multirow{3}{*}{RMLP} &Train. Time (ms) & 2.863 & 3.377 & 3.358 & 3.370 & 3.369 & 3.415\\
&Infer. Time (ms) & 0.882 & 0.991 & 0.977 & 0.973 & 0.963 & 0.992 \\
&Num. of Parameters & 458,158 & 571,670 & 685,189 & 798,806 & 912,521 & 1026,334 \\
\bottomrule
\end{tabular}
\end{table}

\section{Source Code}

 The code associated with this research can be accessed at \url{https://github.com/RogerNi/MoLE}. 

\vfill

\end{document}